\RequirePackage{fix-cm}
\documentclass[smallextended]{svjour3}       
\smartqed  
\usepackage{graphicx}

\usepackage{url}
\usepackage{xcolor}
\usepackage{multirow}
\usepackage{epsfig}
\usepackage{subcaption}
\usepackage{calc}
\usepackage{multicol}
\usepackage{pslatex}
\usepackage{apalike}
\usepackage{lipsum}
\usepackage{subcaption}

\begin{document}

\title{Affine Correspondences in Stereo Vision: Theory, Practice, and Limitations
}


\author{Levente Hajder}


\institute{L. Hajder \at Faculty of Informatics, E\"otv\"os Lor\'and University, P\'azm\'any P\'eter stny. 1/C, Budapest, H-1117, Hungary
              \email{hajder@inf.elte.hu}           
}

\date{Received: date / Accepted: date}

\maketitle

\begin{abstract}
Affine transformations have been recently used for stereo vision. They can be exploited in various computer vision application, e.g., when estimating surface normals, homographies, fundamental and essential matrices. Even full 3D reconstruction can be obtained by using affine correspondences. First, this paper overviews the fundamental statements for affine transformations and epipolar geometry. Then it is investigated how the transformation accuracy influences the quality of the 3D reconstruction. Besides, we propose novel techniques for estimating the local affine transformation from corresponding image directions; moreover, the fundamental matrix, related to the processed image pair, can also be exploited. Both synthetic and real quantitative evaluations are implemented based on the accuracy of the reconstructed surface normals. For the latter one, a special object, containing three perpendicular planes with chessboard patterns, is constructed. The quantitative evaluations are based on the accuracy of the reconstructed surface normals and it is concluded that the estimation accuracy is around a few degrees for realistic test cases. Special stereo poses and plane orientations are also evaluated in detail.
\end{abstract}

\section{Introduction}

\begin{figure}[!t]
\centering
\includegraphics[scale=.4]{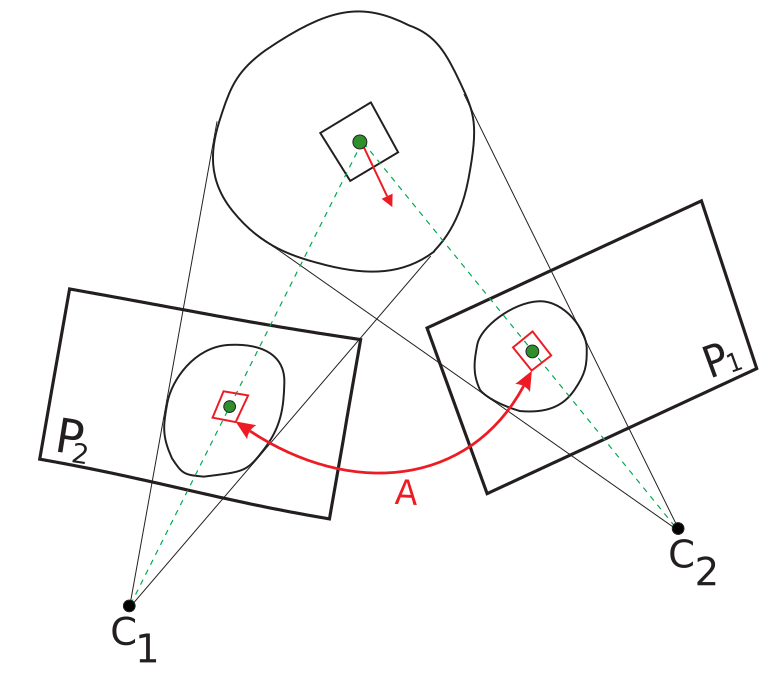}
\caption{Two images of a spatial scene. A small planar surface is perspectively projected to images $P_1$ and $P_2$. The shape deformation between projected patches are transformed by affine transformation $\mathbf A$. \textbf{Green}: Point cloud can be reconstructed from PCs. \textbf{Green+Red}: Oriented point cloud obtained from ACs}
\label{fig:stereo_problem}
\end{figure}

The fundamental goal of geometry-based 3D computer vision \cite{Hartley2003} is to extract spatial information from image. This paper concentrates only on the stereo problem, when 3D data is computed from two images. The geometric problem itself is visualized in Fig.~\ref{fig:stereo_problem}. The coordinates of a spatial point, painted by green color, can be calculated via the so-called triangulation method \cite{HartleyS97} if the corresponding projected locations are known in both images. The camera intrinsic and extrinsic parameters are known as well, in other words, the cameras are calibrated. The intrinsic parameters are retrieved by a calibration method, the most frequently-used one is the chessboard-based method of \cite{Zhang2000}. The stereo geometry is represented by the fundamental and essential matrices for the uncalibrated and calibrated cases, respectively. The relative rotation and translation between the images, i.e. pose of the stereo setup, are obtained by the decomposition of the essential matrix.

This is the standard way to retrieve 3D information from a stereo image pair. Point correspondences (PCs) in the images are required for computing the point cloud of the processed 3D scene. In this paper, this is called \textit{PC-based stereo vision}.

Recently, researchers have begun to deal with processing Affine Correspondences (ACs). If they are exploited for geometric 3D vision problem, it is called \textit{AC-based stereo vision} here. The problem is highlighted in red in Fig.~\ref{fig:stereo_problem}. Two patches are cropped around the corresponding point locations in the images, and the shape deformation between them are represented by an affine transformation $\mathbf A$. In this case, the surface normals can also be estimated as it is written first in the pioneering work of \cite{Molnar2014}.

We think that \textit{this is a novel technology in 3D Stereo Vision}. For PC-based reconstruction, the results are \textit{point clouds}. In AC-based one, spatial points with surface normals can be reconstructed. Accordingly, the output are \textit{oriented points clouds} \cite{RaposoB20}. 

\subsection{Literature Overview}

There are earlier papers that deal with affine transformations: maybe the first study on this field is the paper of \cite{Megyesi2006} that proposes a novel normal estimation method for rectified images in order compute dense 3D model of a scene. \cite{Koser2008} showed that 3D points can be accurately triangulated from local affinities, and surface normals can also be calculated for special stereo setups. \cite{Bentolila2014} proved that an affine transformation gives constraints for estimating the epipoles in the images. 

Since 2014, the following affine transformation-related stereo problems have been solved: (i) \cite{BarathMH15} showed how the normal can be estimated from a calibrated stereo image pair. (ii) They also discussed how homographies and affine transformations relate. Moreover, they proposed methods for estimating homographies both for the cases when the fundamental matrix is or is not known (~\cite{barath2016novel}). (iii) \cite{Raposo016} showed that an affine transformation gives two constraints for the estimation of an essential matrix. A few months later, ~\cite{BarathMH16,BarathTH17} proposed similar equations for the fundamental matrix estimation. Besides, they gave geometric interpretation by showing how the normals of epipolar line pairs are constrained by a corresponding affine transformation.

Affine transformations can be used for other interesting estimation problems such as pose estimation from minimal samples \cite{Ventura2020} including the special planar motion (~\cite{Guan0LSF20,Hajder2020ICRAa,Hajder2020ICRAb}). Additionally, the bundle-adjustment technique can also be applied to AC-based multiple-view 3D reconstruction (\cite{HajderIvan2017}). Recently, AC-based methods have been extended to multiple views \cite{BarathEH19,RaposoB20}. 

This paper focuses on the pin-hole camera model, however, affine transformations can also used for other types of camera lenses \cite{EichhardtC18}. The pioneer on this field was Jozsef Molnar who proposed the general relationship for affine transformations and stereo geometry (\cite{Molnar2014,MolnarE18}). Especially the latter article is a real treasure mine for scientists: the author of this paper believes that several methods will be constructed based on their theoretical statements in the near future.

Even though several geometric problems has been recently solved by AC-based methods, the accuracy of the affine transformations between images is not suitable. This is beacuse feature matcher algorithms~\cite{SIFT2004,Morel2009,SURF2008,MikolajczykTSZMSKG05} focus on the estimation of point correspondences; as a side effect, affine transformations can also be retrieved, however, their qualities are very questionable. Therefore, we think that one of the main goal for high-quality AC-base reconstruction is to develop accurate affine transformation estimator(s).

\subsection{Objectives}

The goal of this paper is threefold: (i) the most important theoretical statements for AC-based stereo vision are overviewed; (ii) the sensitivity of the resulting oriented point cloud is examined w.r.t. quality of affine transformations. Written in the previous paragraph, one of the main future work is to develop novel region matchers that yield accurate ACs. We wish to validated here that accurate surface normals can be reconstructed if the noise in affine transformations is realistic. (iii) We think that the most accurate way to estimate ACs is to use both point and line matches between stereo images. Another important aim of this paper is to propose novel methods that can estimate ACs from locations and directions.

The main contributions of this paper are as follows: (i) We introduce novel algorithms that estimate the affine transformations from scaled or unscaled directions\footnote{Source code of estimators, written in MATLAB, will be available after publication.}.  It is also demonstrated how a fundamental matrix yields two constraints for affinity estimation. (ii) Then a novel 3D reconstruction pipeline is proposed for the validation process that utilizes affine transformations. (iii) The 3D reconstruction results are evaluated on both synthetic data and real images of perpendicular planes with chessboard patterns. (iv) Special planes and stereo poses are also considered.

The structure of this paper is as follows. After this introduction, the geometric background it discussed in Section~\ref{sec:theoretical} including the basic formulas for the relationship of affine transformations and epipolar geometry. Then novel affine transformation estimators are introduced in Sec.~\ref{sec:estimation}. The developed reconstruction pipeline, yielding oriented 3D point clouds, is overviewed in Sec.~\ref{sec:pipeline}. The 3D reconstruction results are evaluated in Sec.~\ref{sec:evaluation}. Finally, Sec.~\ref{sec:conclusion} concludes the work and the results and gives further research directions.

\section{Theoretical Background}
\label{sec:theoretical}




In this section, the most important affine transformation related formulas are overviewed in short: (i) The relationship of an affine transformation, the corresponding surface normal, the spatial location and the camera parameters are discussed first. (ii) Then it is shown how the affine transformation can be obtained from the corresponding homography at a given location. (iii) Finally, the relation of corresponding epipolar lines and affine transformations are discussed.

\subsection{Relationship of Camera Parameters, a Surface Normal and Related Affine Transformation}
\label{sec:formula_normal}
An affine transformation is represented by a $2 \times 2$ matrix denoted by
\begin{equation}
\label{eq:A}
    \mathbf A=\left[ \begin{array}{cc} a_1 &a_2 \\ a_3 & a_4 \end{array} \right].
\end{equation}
As it was derived by \cite{Molnar2014}, an affine transformation depends on the related surface normals, the spatial location, and the camera intrinsic and extrinsic parameters. The related formula is as follows:
\begin{equation}
 \label{eq:general_aff_proj}
\mathbf A = \frac{1}{\mathbf n^T \mathbf w_5}\left[  \begin{array}{cc} \mathbf n^T \mathbf w_1 & \mathbf n^T \mathbf w_2 \\ \mathbf n^T \mathbf w_3 & \mathbf n^T \mathbf w_4 \end{array} \right],
\end{equation}
where 
\begin{equation}
\begin{array}{c}
\mathbf w_1= \left[\frac{\partial v_1}{\partial X}, \frac{\partial v_1}{\partial X} , \frac{\partial v_1}{\partial Z}  \right]^T \times \left[\frac{\partial u_1}{\partial X}, \frac{\partial u_1}{\partial X} , \frac{\partial u_1}{\partial Z}  \right]^T, \\
\mathbf w_2= \left[\frac{\partial u_2}{\partial X}, \frac{\partial u_2}{\partial X} , \frac{\partial u_2}{\partial Z}  \right]^T \times \left[\frac{\partial u_1}{\partial X}, \frac{\partial u_1}{\partial X} , \frac{\partial u_1}{\partial Z}  \right]^T, \\
\mathbf w_3= \left[\frac{\partial v_1}{\partial X}, \frac{\partial v_1}{\partial X} , \frac{\partial v_1}{\partial Z}  \right]^T \times \left[\frac{\partial v_2}{\partial X}, \frac{\partial v_2}{\partial X} , \frac{\partial v_2}{\partial Z}  \right]^T,\\
\mathbf w_4= \left[\frac{\partial v_2}{\partial X}, \frac{\partial v_2}{\partial X} , \frac{\partial v_2}{\partial Z}  \right]^T \times \left[\frac{\partial u_1}{\partial X}, \frac{\partial u_1}{\partial X} , \frac{\partial u_1}{\partial Z}  \right]^T,\\
\mathbf w_5= \left[\frac{\partial v_1}{\partial X}, \frac{\partial v_1}{\partial X} , \frac{\partial v_1}{\partial Z}  \right]^T \times \left[\frac{\partial u_1}{\partial X}, \frac{\partial u_1}{\partial X} , \frac{\partial u_1}{\partial Z}  \right]^T,
\end{array}
\label{eq:ws}
\end{equation}
if $[u_1 , v_1]^T$ and $[u_2 , v_2]^T$ are the 2D coordinates of the centers of the projected patches in the images, while $[X,Y,Z]^T$ is the spatial point.
Equation~\ref{eq:general_aff_proj} is a general formula that connects the affine transformations and camera parameters. Only the gradients of the projective functions are calculated that depend on the applied camera model. 

\noindent \textbf{For a pin-hole camera}, the projection is written as

\begin{equation}
\left[ \begin{array}{c} u_i \\ v_i \\ 1 \end{array}  \right]  \sim \left[ \begin{array}{cccc} p^i_1 & p^i_2 & p^i_3 & p^i_4 \\ p^i_5 & p^i_6 & p^i_7 & p^i_8 \\ p^i_9 & p^i_{10} & p^i_{11} & p^i_{12}  \end{array} \right] \left[ \begin{array}{c} X \\ Y \\ Z \\ 1 \end{array}  \right] ,
\end{equation}

\noindent or, by coordinates, after homogeneous division:

\begin{equation}
\begin{array}{c}
u_i = \frac{p^i_1 X + p^i_2 Y + p^i_3 Z + p^i_4}{p^i_9 X + p^i_{10} Y + p^i_{11} Z + p^i_{12}} , \\
v_i = \frac{p^i_5 X + p^i_6 Y + p^i_7 Z + p^i_8}{p^i_9 X + p^i_{10} Y + p^i_{11} Z + p^i_{12}}.
\end{array}
\end{equation}

\noindent The derivatives are as follows:

\begin{equation}
\begin{array}{cc}
\frac{\partial u_i}{\partial X} =\frac{1}{q} \left( p^i_{1} + u_i p^i_{9}    \right),   &
\frac{\partial u_i}{\partial Y} = \frac{1}{q} \left( p^i_{2} + u_i p^i_{10}    \right), \\
\frac{\partial u_i}{\partial Z} = \frac{1}{q} \left( p^i_{3} + u_i p^i_{11} \right), &
\frac{\partial v_i}{\partial X} = \frac{1}{q} \left( p^i_{5} + v_i p^i_{9}  \right), \\
\frac{\partial v_i}{\partial Y} = \frac{1}{q} \left( p^i_{6} + v_i p^i_{10} \right),  &
\frac{\partial v_i}{\partial Z} = \frac{1}{q} \left( p^i_{7} + v_i p^i_{11} \right),  
\end{array}
\end{equation}
where $q=p^i_9 X + p^i_{10} Y + p^i_{11} Z + p^i_{12}$. These values are substituted back to Eqs.~\ref{eq:ws}, then the obtained vectors in Eq.~\ref{eq:general_aff_proj} can be determined in order to obtain the relationship of affine parameters, normals, camera parameters and spatial points for the pin-hole camera model.

\subsection{Homography and Affine Transformations}
\label{sec:formula_homography}
If a planar surflet is observed by two images, the transformation between the images is represented by the so-called homography. Formally, it can be written as

\begin{equation}
    \left[ \begin{array}{c} u_2 \\ v_2 \\ 1 \end{array} \right] \sim \mathbf H  \left[ \begin{array}{c} u_1 \\ v_1 \\ 1 \end{array} \right]= \left[ \begin{array}{ccc} h_1 & h_2 & h_3 \\ h_4 & h_5 & h_6 \\ h_7 & h_8 & h_9 \end{array} \right]  \left[ \begin{array}{c} u_1 \\ v_1 \\ 1 \end{array} \right],
\end{equation}

\noindent or by coordinates, after homogeneous division:

\begin{equation}
\begin{array}{cc}
u_2 = \frac{h_1 u_1 + h_2 v_1 + h_3}{h_7 u_1 + h_8 v_1 + h_9}, &
v_2 = \frac{h_4 u_1 + h_5 v_1 + h_6}{h_7 u_1 + h_8 v_1 + h_9}.
\end{array}
\end{equation}

The affine transformation is the first-order derivative of the transformation w.r.t. directions in image space. Thus, its elements are

\begin{equation}
\begin{array}{cc}
a_1 = \frac{\partial u_2}{ \partial u_1}=\frac{h_1 - h_7 u_2}{s},
&
a_2 = \frac{\partial u_2}{ \partial v_1}=\frac{h_2 - h_8 v_2}{s},
\\
a_3 = \frac{\partial v_2}{ \partial u_1}=\frac{h_4 - h_7 u_2}{s},
&
a_4 = \frac{\partial v_2}{ \partial v_1}=\frac{h_5 - h_8 v_2}{s},
\end{array}
\end{equation}
where $s=h_7 u_1 + h_8 v_1 + h_9$. These four equations give the relationship of the elements of an affine transformation and the related homography. It is well-known \cite{Hartley2003} that a point correspondence gives two equations, therefore, one AC yields four equations in total. As a result, only two ACs are required for the eight degrees of freedom (DoFs) of a homography. Moreover, the estimation problem is over-determined in this case.

Remark, that is also shown in \cite{BarathH17} that single AC is enough if the fundamental matrix is known as the fundamental matrix reduces the DoF of the homography estimation by six. It is in conjunction with the previous section (Sec~\ref{sec:formula_homography}): a 3D point and the corresponding surface normal, that can be estimated from one AC if the cameras are calibrated, exactly define a plane in the 3D space. For a calibrated camera pair, a homography can be decomposed into a 3D point and a normal \cite{Faugeras1988}; and the homography itself can be estimated from a single AC if the epipolar geometry is known.

\subsection{Fundamental or Essential Matrices and Affine Transformation}
\label{sec:formula_fundamental}
\begin{figure}[!t]
\centering
\includegraphics[width=\textwidth]{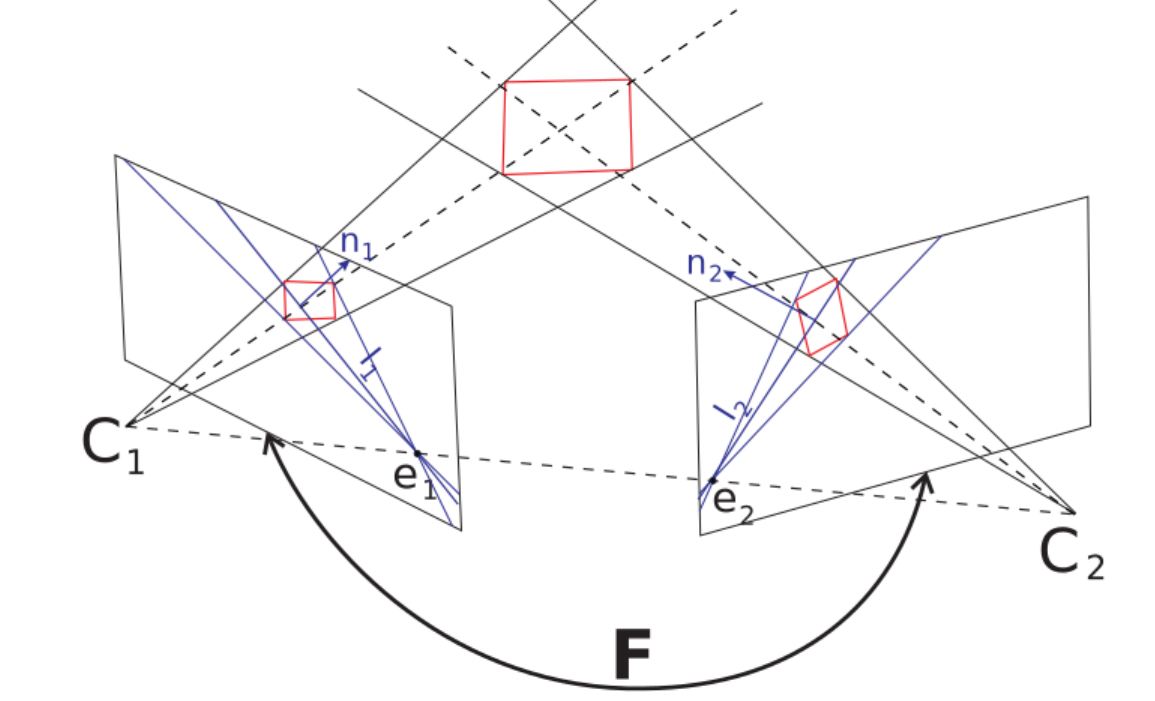}
\caption{An affine transformation transforms the shape of the red quadrilaterals from the first to the second images. It also connects normals of corresponding epipolar lines. The direction and scale changes of normals $\mathbf n_1$ and $\mathbf n_2$ are determined by the affine transformation. $\mathbf C_1$, $\mathbf C_2$, $\mathbf e_1$, and $\mathbf e_2$ are focal points and epipoles, respectively.}
\label{fig:fund_affine}
\end{figure}

The relationship between point locations, affine transformation, and a fundamental matrix is given in the paper of \cite{BarathTH17}. The setup is visualized in Fig.~\ref{fig:fund_affine}. It can be written via the normals of the corresponding epipolar lines. It is as follows:
\begin{equation}
\label{eq:basic_A_F}
\mathbf{A}^{T} \mathbf n_2 =- \mathbf n_1,
\end{equation}
where
normals of the corresponding epipolar lines are denoted by 2D vectors $\mathbf n_1$ and $\mathbf n_2$. The epipolar lines in the first and second images are calculated as
\begin{equation}
\label{eq:epip_lines}
    \left[ \begin{array}{c} l_{2u} \\ l_{2v} \\ l_{2w} \end{array} \right] = \mathbf{F} \left[ \begin{array}{c} x_1 \\ y_1 \\ 1 \end{array} \right], \quad 
        \left[ \begin{array}{c} l_{1u} \\ l_{1v} \\ l_{1w} \end{array} \right] = \mathbf{F^T} \left[ \begin{array}{c} x_2 \\ y_2 \\ 1 \end{array} \right],
\end{equation}
and the normals, applied in Eq.~\ref{eq:basic_A_F}, are $\mathbf n_1 = [l_{1u},l_{1v}]^T$ and $\mathbf n_2 = [l_{2u},l_{2v}]^T$. The length of the normals are used in the estimation procedure, thus, they are not normalized.

\section{Estimation of Affine Transformations from Corresponding Directions}
\label{sec:estimation}

The objective of this section is to estimate an affine transformation corresponding to a stereo point correspondence in the images and corresponding directions. We introduce several algorithms here. The preliminary version of the algorithms published in \cite{MinhH20}, however, novel methods are also proposed here. Remark that the source code of the algorithms will be published.

The locations in the images are denoted by $\mathbf{p_1}=\left[\begin{array}{cc}
u_1 & v_1\end{array}\right]^{T}$ and  $\mathbf{p_2}=\left[\begin{array}{cc}
u_2 & v_2\end{array}\right]^{T}$, while the directions by $ \mathbf{d}_{1i}=\left[\begin{array}{cc}
u_{1i} & v_{1i}\end{array}\right]^{T}$ and $\mathbf{d}_{2i}=\left[\begin{array}{cc}
u_{2i} & v_{2i}\end{array}\right]^{T},i\in\left\{ 1,2,\dots,N\right\}$. 
The affine transformation transform the directions from the first image to the second one, however, the lengths are not always known. Therefore the relation can be only written by introducing a scale variable $\alpha_i$. Then the formula becomes $\alpha_i \mathbf d_{i2}=\mathbf{A} \mathbf d_{i2}$ in short, or 
\begin{equation}
    \alpha_i \left[\begin{array}{c}
u_{i2}\\
v_{i2}
\end{array}\right]=\left[\begin{array}{cc}
a_{1} & a_{2}\\
a_{3} & a_{4}
\end{array}\right]\left[\begin{array}{c}
u_{i1}\\
v_{i1}
\end{array}\right],
\end{equation}
if the matrix and vector elements are expressed. The formula can be rewritten as
\begin{equation}
\label{eq:scaled_dirs_lin}
\left[\begin{array}{cccc}
u_{i1} & v_{i1} & 0 & 0\\
0 & 0 & u_{i1} & v_{i1}
\end{array}\right]\left[\begin{array}{c}
a_{1}\\
a_{2}\\
a_{3}\\
a_{4}
\end{array}\right]=\alpha_{i}\left[\begin{array}{c}
u_{i2}\\
v_{i2}
\end{array}\right].
\end{equation}
This is an inhomogeneous linear system of equations w.r.t. elements of the affine transformation.

If the scale factor $\alpha_i$ is unknown, it should be moved into the vector of unknown variables, then a homogeneous linear problem is obtained as follows:
\begin{equation}
\label{eq:unscaled_dirs_lin}
\left[\begin{array}{ccccc}
u_{i1} & v_{i1} & 0 & 0 & -u_{i2}\\
0 & 0 & u_{i1} & v_{i1} & -v_{i2}
\end{array}\right]\left[\begin{array}{c}
a_{1}\\
a_{2}\\
a_{3}\\
a_{4}\\
\alpha_{i}
\end{array}\right]=\mathbf 0.
\end{equation}
This case is called the unslaced one. If $\alpha_i$ is known, it is called the scaled case.

In summary, two linear equations can be written to both scaled and unscaled directions. The former one, defined by Eq.~\ref{eq:scaled_dirs_lin}, is an inhomogoeneous system, while the homogoeneous system of Eq.~\ref{eq:unscaled_dirs_lin} represents the unscaled case.

\subsection{Estimation of an Affine Transformation from Unscaled Directions}
If three unscaled directions are available, the following homogeneous linear system, based on Eq.~\ref{eq:unscaled_dirs_lin}, can be written:

\begin{equation}
\label{eq:three_unscaled_dirs}
\left[\begin{array}{ccccccc}
u_{11} & v_{11} & 0 & 0 & -u_{12} & 0 & 0\\
0 & 0 & u_{11} & v_{11} & -v_{12} & 0 & 0\\
u_{21} & v_{21} & 0 & 0 & 0 & -u_{22} & 0\\
0 & 0 & u_{21} & v_{21} & 0 & -v_{22} & 0\\
u_{31} & v_{31} & 0 & 0 & 0 & 0 & -u_{32}\\
0 & 0 & u_{31} & v_{31} & 0 & 0 & -v_{32}
\end{array}\right]\left[\begin{array}{c}
a_{1}\\
a_{2}\\
a_{3}\\
a_{4}\\
\alpha_{1}\\
\alpha_{2}\\
\alpha_{3}
\end{array}\right]=\mathbf 0
\end{equation}
In matrix-vector form, it can be written as $\mathbf{B} \mathbf{x}=\mathbf 0$, where $\mathbf{x}=\left[ a_1,\dots,a_4,\alpha_1, \dots ,\alpha_3 \right]^T$, and $\mathbf B$ is the left matrix of Eq.~\ref{eq:three_unscaled_dirs}. As the problem is homogeneous, the solution can only be obtained up to an unknown scale. This scale is an a-priori information for the algorithm. Typically, the affine transformation modifies the size of the patches, which gives the scale. For an affine transformation, the area change is equivalent to the determinant of the affine transformation itself. If there is no a-priori information, an acceptable approximation is that there is no size change, thus the determinant is the unit: $det \left( \mathbf{A} \right)=a_1 a_4 - a_2 a_3=1$.

\noindent \textbf{Minimal solution.} In this case, the coefficient matrix $\mathbf B$ generally has a single null-vector. The scale of the vector can be determined if one constraint is given. 

\noindent \textbf{Over-determined problem.}
If $N$ directions are known, and $N>3$ the coefficient matrix has $2N$ rows, the problem is over-determined. In this case, there is no null-vector. The solution is obtained by a generalized eigenvector problem as it is discussed in Appendix~\ref{sec:app}. The scale must be known for the algorithm as Eq.~\ref{eq:three_unscaled_dirs} does not contain any information about it. 

\subsection{Estimation of Affine Transformation from Two Scaled Directions}

The affine estimation problem can be exactly solved if two scaled directions are known. In this case, the formula in Eq.~\ref{eq:scaled_dirs_lin} can be applied, and the estimation becomes an inhomogeneous linear system of four equations as follows:

\begin{equation}
\left[\begin{array}{cccc}
u_{11} & v_{11} & 0 & 0\\
0 & 0 & u_{11} & v_{11}\\
u_{21} & v_{21} & 0 & 0\\
0 & 0 & u_{21} & v_{21}
\end{array}\right]\left[\begin{array}{c}
a_{1}\\
a_{2}\\
a_{3}\\
a_{4}
\end{array}\right]=\left[\begin{array}{c}
\alpha_{1}u_{12}\\
\alpha_{1}v_{12}\\
\alpha_{2}u_{22}\\
\alpha_{2}v_{22}
\end{array}\right].
\end{equation}
Scale variables $\alpha_1$ and $\alpha_2$ are known, therefore, only the elements of the affine transformation $\mathbf A$ should be estimated.

\subsection{Estimation of an Affine Transformation if Epipolar Geometry is Known}
\label{sec:affine_estimation}

The goal of this section is to show how an affinity can be estimated if the fundamental matrix is known for the stereo image pair. Additionally, a corresponding point pair and two directions should be given.
The relationship between point locations, affine transformation, and the fundamental matrix is given by Eqs.~\ref{eq:basic_A_F} and~\ref{eq:epip_lines}. If the elements of the affine transformations are considered, and the coordinates of the normals are substituted, the following linear system of equations is obtained from Eq.~\ref{eq:basic_A_F}:

\begin{equation}
\label{eq_affine1}
\left[\begin{array}{cc}
a_{1} & a_{3}\\
a_{2} & a_{4}
\end{array}\right]\left[\begin{array}{c}
l_{2u}\\
l_{2v}
\end{array}\right]=-\left[\begin{array}{c}
l_{1u}\\
l_{1v}
\end{array}\right].
\end{equation}

If two additional directions are known in the images, that are connected by the affine transformations, the problem becomes solvable. Therefore Eq.~\ref{eq:unscaled_dirs_lin} should be applied twice for the unscaled directions. The final linear system is as follows:

\begin{eqnarray*}
\left[\begin{array}{cccccc}
l_{2u} & 0 & l_{2v} & 0 & 0 & 0\\
0 & l_{2u} & 0 & l_{2v} & 0 & 0\\
u_{11} & v_{11} & 0 & 0 & -u_{21} & 0\\
0 & 0 & u_{11} & v_{11} & -v_{21} & 0\\
u_{12} & v_{12} & 0 & 0 & 0 & -u_{22}\\
0 & 0 & u_{12} & v_{12} & 0 & -v_{22}
\end{array}\right]\left[\begin{array}{c}
a_{1}\\
a_{2}\\
a_{3}\\
a_{4}\\
\alpha_{1}\\
\alpha_{2}
\end{array}\right]= 
\left[\begin{array}{c}
-l_{1u}\\
-l_{1v}\\
0\\
0\\
0\\
0
\end{array}\right].
\end{eqnarray*}

\section{Reconstruction Pipeline}
\label{sec:pipeline}
For the testing of the algorithms, a novel reconstruction pipeline is developed. The main goal of this pipeline is to include the affine transformation, estimated by the method proposed in Sec.~\ref{sec:estimation}, into the 3D reconstruction. Thus, the final 3D model of the scene is represented by an oriented point cloud for which the orientation is given by the estimated surface normals.

\begin{figure}[!t]
\centering
\includegraphics[scale=0.33]{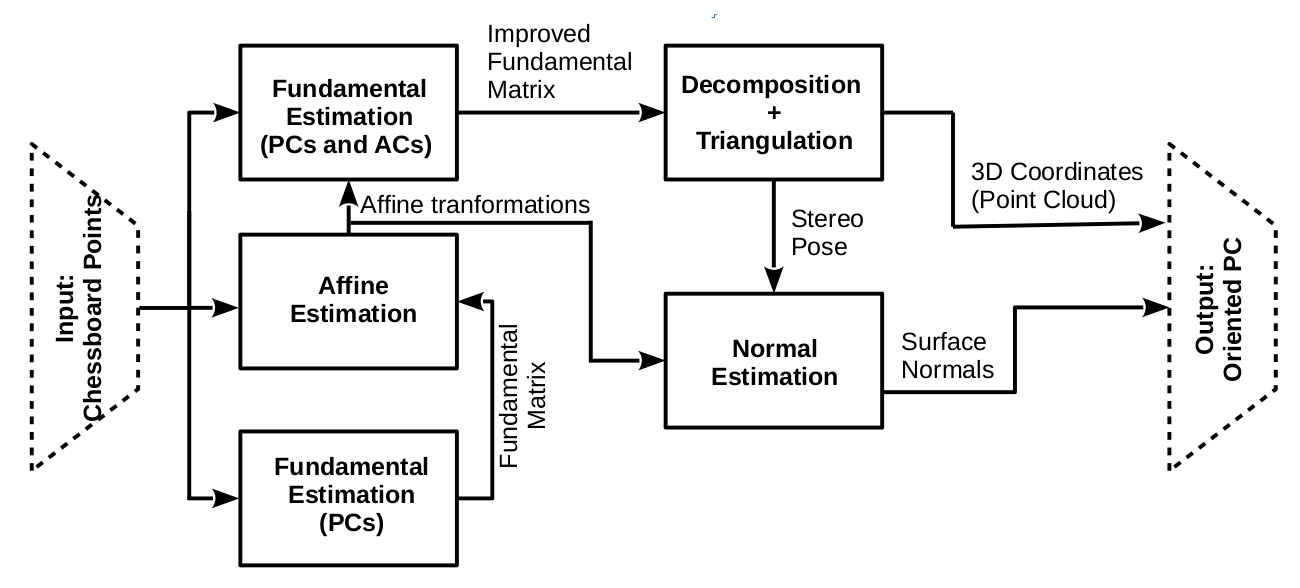}
\caption{Block diagram of proposed pipeline for AC-based 3D reconstruction. This pipeline is designed for the quantitative evaluation. Its input are only the chessboard corners. The output are the reconstructed oriented point clouds: spatial locations and normals of reconstructed surfaces.}
\label{fig:pipeline}
\end{figure}

The pipeline is overviewed in Fig~\ref{fig:pipeline}. It is similar to the one proposed earlier by \cite{HajderIvan2017}, however, there are significant differences: (i) The main goal here is to construct a pipeline for the validation, therefore, it is not a general pipeline. However, if the directions for 'Affine Estimation' are given as input data, the pipeline can be straightforwardly generalized.  (ii) Affine transformations are detected by the novel methods, proposed in this paper; (iii) ACs are used to estimate the epipolar geometry contrast to~\cite{HajderIvan2017} where only PCs are applied. 


The \textit{input} for the pipeline are point correspondences in images, located at chessboard corners, and pre-calibrated camera parameters. The \textit{output} is the reconstructed oriented point cloud.

The steps of the pipeline are as follows:
\begin{itemize}
    \item \textbf{Chessboard Corner Estimation.} Three perpendicular chessboards are seen in the input images. Their corners are the input of our pipeline. The detector of \cite{Geiger2012ICRA} is applied for chessboard corner localization as it can detect corners of multiple chessboard planes.
    \item \textbf{Fundamental matrix estimation from PCs.} The standard normalized-eight point algorithm \cite{Hartley1997defence} is applied. The fundamental matrix $\mathbf F$ is refined later from both PCs and ACs, however, an initial value is required if affine transformations are estimated exploiting the fundamental matrix.
    \item \textbf{Affine Estimation from PCs + ACs.} The affine transformations are estimated by the proposed methods, discussed in Sec.~\ref{sec:estimation}.
    \item \textbf{Fundamental matrix estimation from PCs and ACs.} The fundamental matrix is re-estimated because affine parameters can improve its accuracy. For general motion, the method of \cite{BarathTH17} is applied. For special motions like planar motion, forward motion, or standard stereo, our planar essential/fundamental matrix estimation method (\cite{Hajder2020ICRAa}) is selected.
    \item \textbf{Decomposition + Triangulation.} The essential matrix represents the pose of the two images, it can be retrieved from the fundamental matrix \cite{Hartley2003} if the cameras are calibrated. Then the essential matrix is decomposed into rotation and translation, and the 3D point cloud is obtained by the triangulation algorithm of \cite{HartleyS97}\footnote{The four candidate solution for rotation and translation are also selected by triangulation.}.
    \item \textbf{Normal Estimation.} Finally, the surface normals are calculated by the optimal method proposed in \cite{barath2016novel}. The solver for surface normals is based on the formulas discussed in Sec.~\ref{sec:formula_normal}.
\end{itemize}

\section{Experiments}
\label{sec:evaluation}

The test section of this paper is very special as the main goal is not to compare the proposed method to the State of the Art (SoA) algorithms but to evaluate the limits of using affine correspondences in a 3D reconstruction pipeline. 

Synthetic and real examples are also considered. For real-world tests, very accurate PCs and ACs should be used. In computer vision, chessboards are frequently used as the intersecting points of the chessboard fields can be efficiently and accurately detected even if non-perspective optics are applied \cite{Geiger2012ICRA,Schoenbein2014ICRA}. As we showed in the paper, local affine transformation can be estimated from corresponding image directions which can be  obtained very accurately exploiting the chessboard corners. The potential directions are visualized in Fig.~\ref{fig:directions}.

\begin{figure}[ht]
\begin{subfigure}{.5\textwidth}
  \centering
\includegraphics[scale=.24]{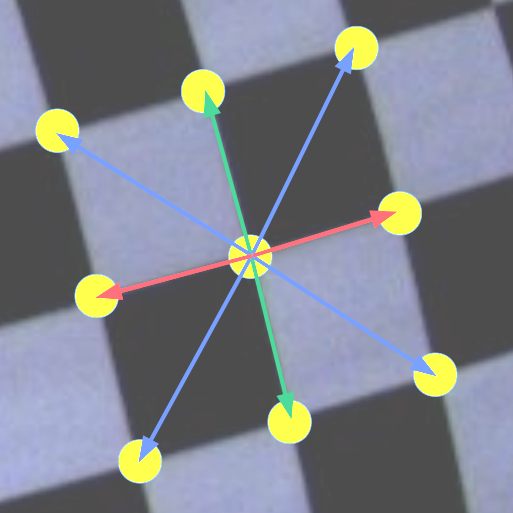}
  \caption{Directions in chessboard pattern.}
\label{fig:directions}
\end{subfigure}
\begin{subfigure}{.5\textwidth}
  \centering
\includegraphics[scale=.2]{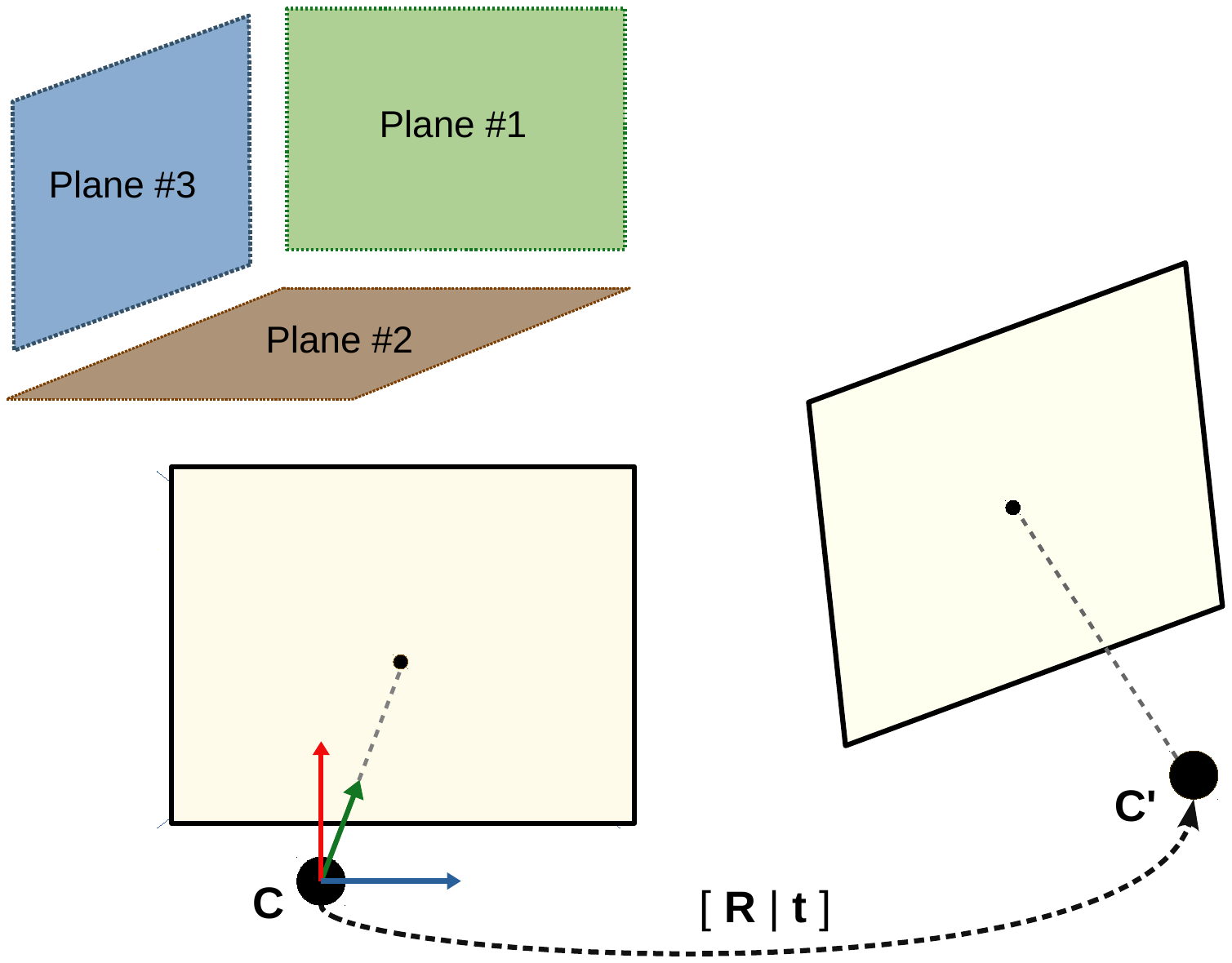}
\caption{Plane numbering for tests.}
\label{fig:synthetic_planes}
\end{subfigure}
\caption{\textbf{Left.} Three different directions are considered, each connects chessboard corners: horizontal (red), vertical (green), diagonal (purple). The horizontal and vertical directions are exploited for affine transformation estimation when two directions are used (methods F2UDIR and 2SDIR); diagonal directions are applied for three direction-based estimation (methods F3UDIR, DET3UDIR, 3SDIR). For the diagonal direction, only one out of the two possible directions is utilized.
\textbf{Right.} The structure of the planes for special motions. Plane \#1, \#2, \#3  are perpendicular to the axes $Z$, $Y$, and $X$, respectively.
}
\label{fig:combined}
\end{figure}


We constructed a 3D object, consisting of three perpendicular chessboards. Four different types of poses are used for the stereo tests. Example images are drawn in Figs.~\ref{fig:real_normal} and~\ref{fig:real_fisheye}. The sequences are labelled by the types of poses between the images. The considered pose types are as follows:
\begin{itemize}
\item \textbf{General motion.} The parameters for the camera pose, including three coordinates of translation vector and three Euler angles, are randomly selected.
\item \textbf{Planar motion.} The motion is represented by two angles: a rotation and a translation\footnote{As only the direction of the translation can be reconstructed, it can be represented by a single angle.}. It is a widely-used model in robotic vision \cite{Ortin2001} or for cameras of autonomous vehicles \cite{Guan0LSF20}. This model is valid for a vehicle-mounted camera if the image plane is perpendicular to the road surface, and the road itself is planar.
\item \textbf{Standard stereo.} For standard stereo, the pose is represented by a single parameter, the baseline, i.e. the length of horizontal translation. The orientations are the same for both cameras.
\item \textbf{Forward Motion.} Similarly to the standard stereo, the orientations are the same. Only the third coordinate of the translation vector is non-zero, thus the translation is parallel to the optical axis.
\end{itemize}

In order to estimate the affine transformations which are then used as an input to the reconstruction pipeline, five different algorithm are applied. The ideas behind those are discussed above in Sec.~\ref{sec:estimation}. The algorithms are as follows:

\begin{itemize}
\item \textbf{F2UDIR.} The affine transformation is estimated from the known fundamental matrix; the horizontal, and vertical directions.
\item \textbf{F3UDIR.} It is similar to F2DIR, however, three directions, including the diagonal one, are considered in the estimation. The problem is over-determined as two unscaled directions are sufficient for the affine estimation if the epipolar geometry is known.
\item \textbf{DET3UDIR.} Only three unscaled directions are used for the estimation. The scale of the affine transformation is given by the three scales of the three possible chessboard directions: horizontal, vertical and diagonal. The squared average of the scales are set as the determinant of the affine transformations.
\item \textbf{2SDIR.} Two scaled directions, the horizontal and the vertical ones, are considered.
\item \textbf{3SDIR.} Three scaled directions, including the diagonal one, are considered, thus, the problem is over-determined.
\end{itemize}


The aim of the tests is to find the limitation of the normal reconstruction w.r.t.  affine parameters. The affine transformations are contaminated by noise. As they are estimated principally from directions, zero-mean Gaussian noise is added to the angles of the directions in the image space. The length of the directional vectors are not changed. The spread of the noise is a parameter, it is called "Directional Error" in the tests.

\subsection{Synthetic Tests}

The synthetic testing environment is implemented in Octave\footnote{http://www.octave.org}. The scenes consist of three perpendicular planes with simulated chessboard corners, similarly to the real test. The dimension of the chessboard corners are $6 \times 8$. The four aforementioned stereo pose types are considered. The chessboard corners in the images are obtained by projecting the 3D points into the generated camera images. For the special cases, the planes are visualized in Fig.~\ref{fig:synthetic_planes}: Plane \#1 is parallel to the first camera image, Plane~ \#2 is the horizontal one, and the vertical Plane \#3 is perpendicular to the other ones.  Finally, the reconstruction pipeline, discussed above, is applied.

\begin{figure}[!t]
\centering
\includegraphics[scale=.35]{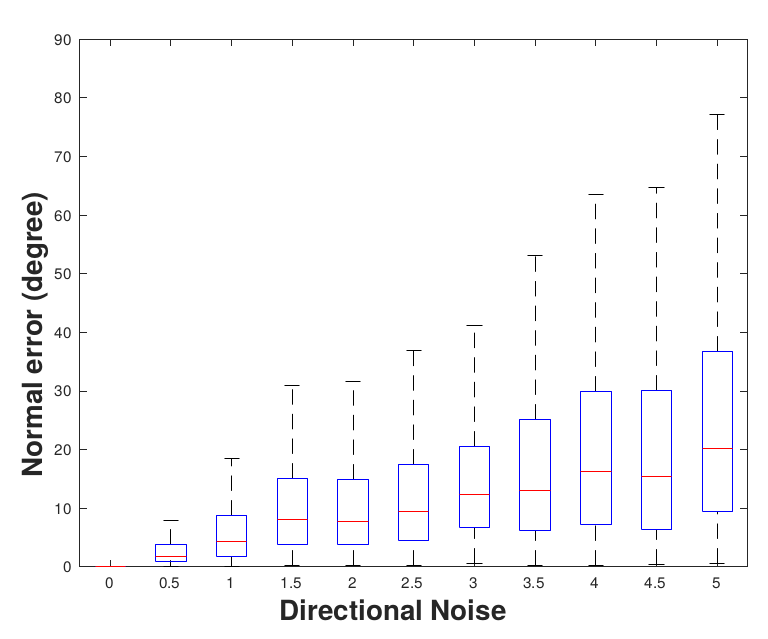}
\caption{Directional error of the normals w.r.t. plane normals. General stereo pose is considered. Directional noise, given in degrees, is added to the angle of 2D directions. The affine transformations are computed by the F2UDIR algorithm. The average error for all planes are plotted.}
\label{fig:synthetic_general}
\end{figure}

To validate the methods, the resulting oriented point clouds are examined. There are three planes, these planes can be reconstructed from the triangulated point cloud by the well-known Principal Component Analysis (PCA) \cite{PCA1986} method. The planes are estimated independently, we ignored the fact that they are orthogonal. PCA yields the normal of the planes. Then the individual normals, generated from the pipeline, are compared to the plane normals, and we used the angular error to measure the accuracy.

In order to visualize the results, boxplot method of MATLAB is called. On each box, the central mark indicates the median, and the bottom and top edges of the box indicate the 25th and 75th percentiles, respectively. The whiskers extend to the most extreme data points.

\begin{figure}[!t]
\centering
\includegraphics[scale=.35]{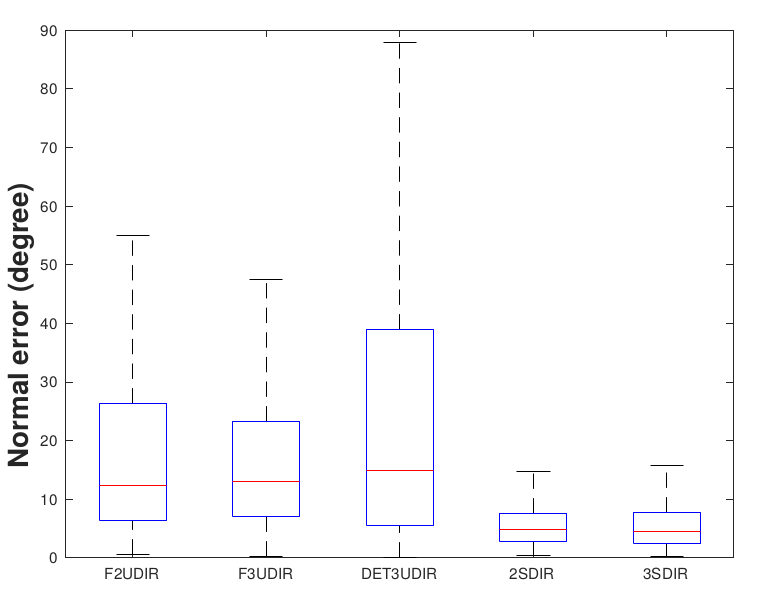}
\caption{Quantitative comparison of the proposed affine estimators in case of synthetic input. General stereo pose is considered. The noise was added to the chessboard directions by a zero-mean Gaussian error with spread $3.0^{\circ}$. The average directional error of the surface normals are computed. The affine transformations are estimated by the proposed algorithms.}
\label{fig:synthetic_AffineModes}
\end{figure}

First, the general motion is considered. Results are visualized in Figs.~\ref{fig:synthetic_general} and \ref{fig:synthetic_AffineModes}. For the former test, the normal error is evaluated w.r.t. directional noise, while the affine estimators are compared for the latter test. It is clearly seen that the quality of surface normals are accurate even if the noise is high. The median error is $20^{\circ}$ at most that is a reasonable quality for high noise.

When the affine estimators are compared, results are plotted in Fig~\ref{fig:synthetic_AffineModes}, the noise level for the directions is set to $3.0^{\circ}$. The conclusion of this tests that it is beneficial if scaled directions are used as the results for methods 2SDIR and 3SDIR are significantly more accurate than those for DET3UDIR. The fundamental matrix based methods (F2UDIR and F3UDIR) are in the middle of the competition. However, the significance of these methods is very high as it is very challenging to retrieve the scales for directions, while the fundamental matrix can be efficiently estimated by standard 3D Vision algorithms \cite{Hartley2003}.

\begin{figure*}[!t]
\centering
\begin{subfigure}{1.0\textwidth}
  \centering
\includegraphics[scale=.27]{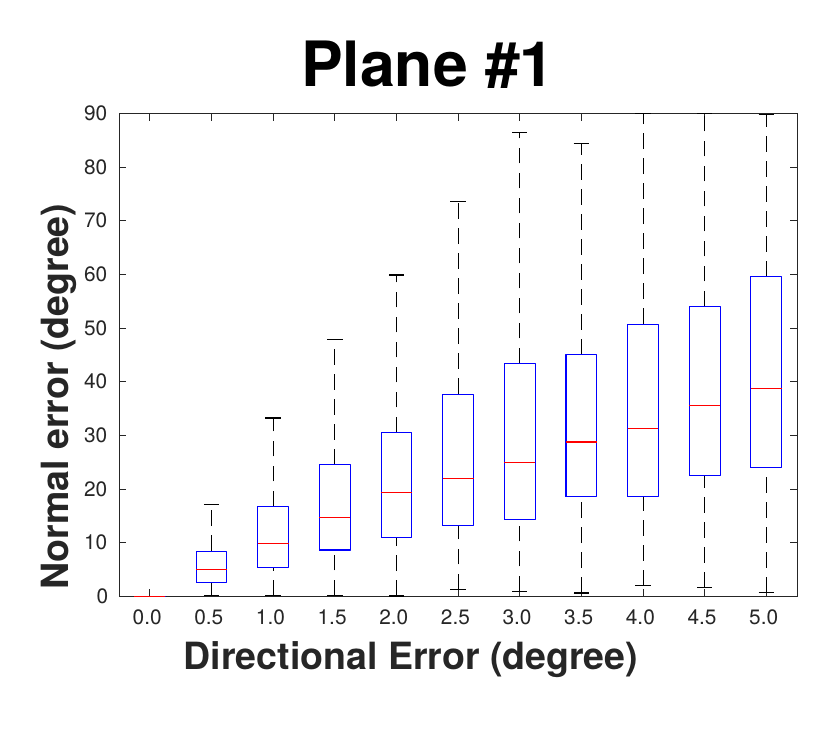}
\includegraphics[scale=.27]{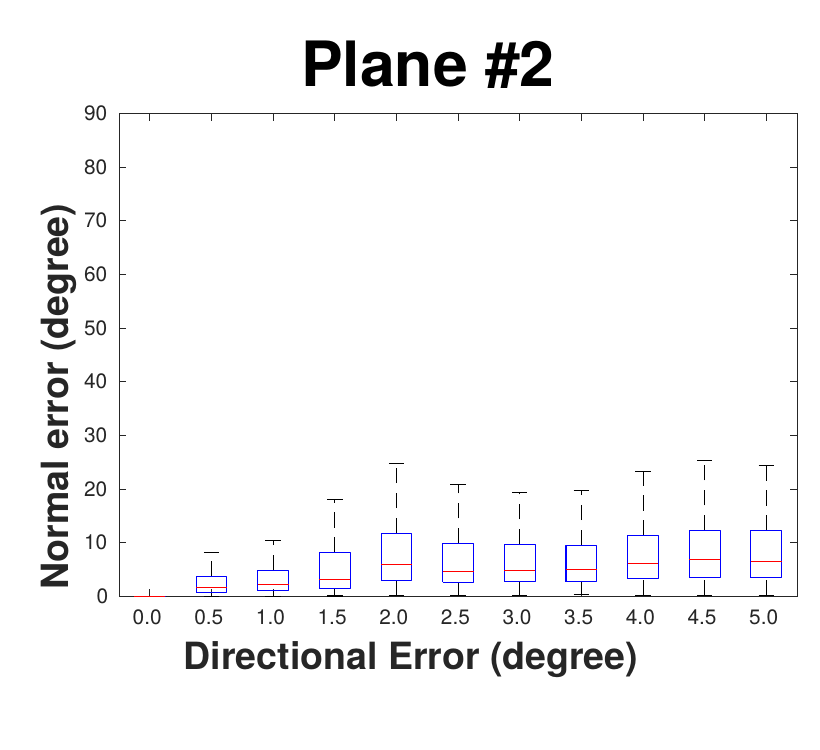}
\includegraphics[scale=.27]{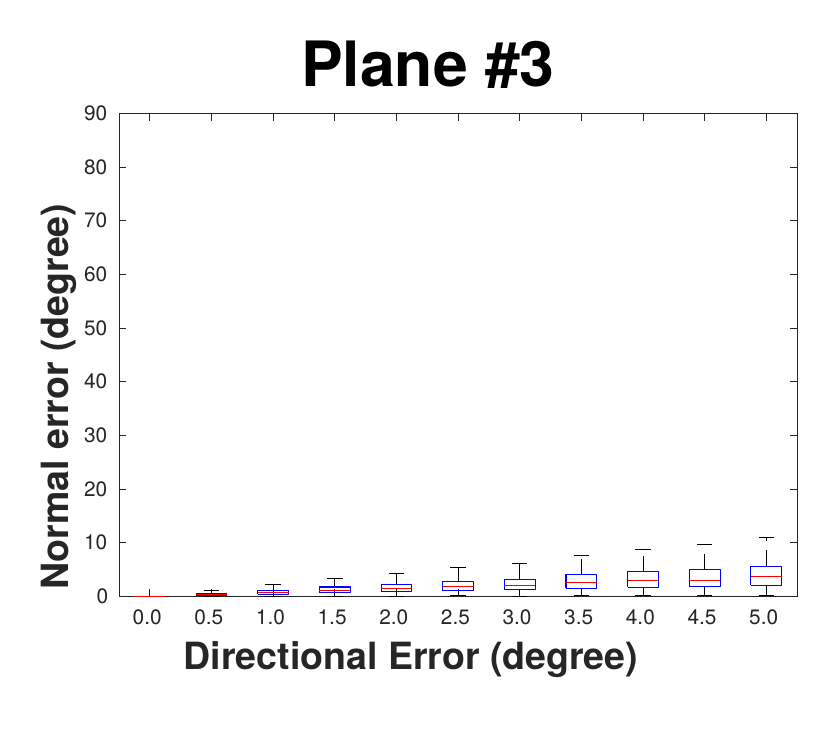}
  \caption{Planar Motion.}
  \label{fig:chart_synthetic_planar}
\end{subfigure}
\begin{subfigure}{1.0\textwidth}
  \centering
\includegraphics[scale=.27]{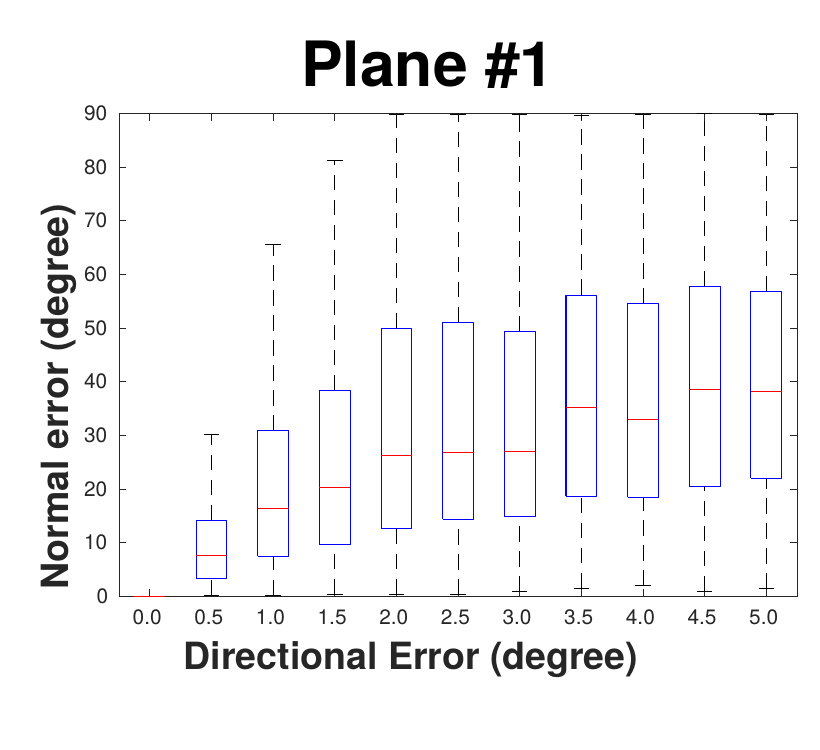}
\includegraphics[scale=.27]{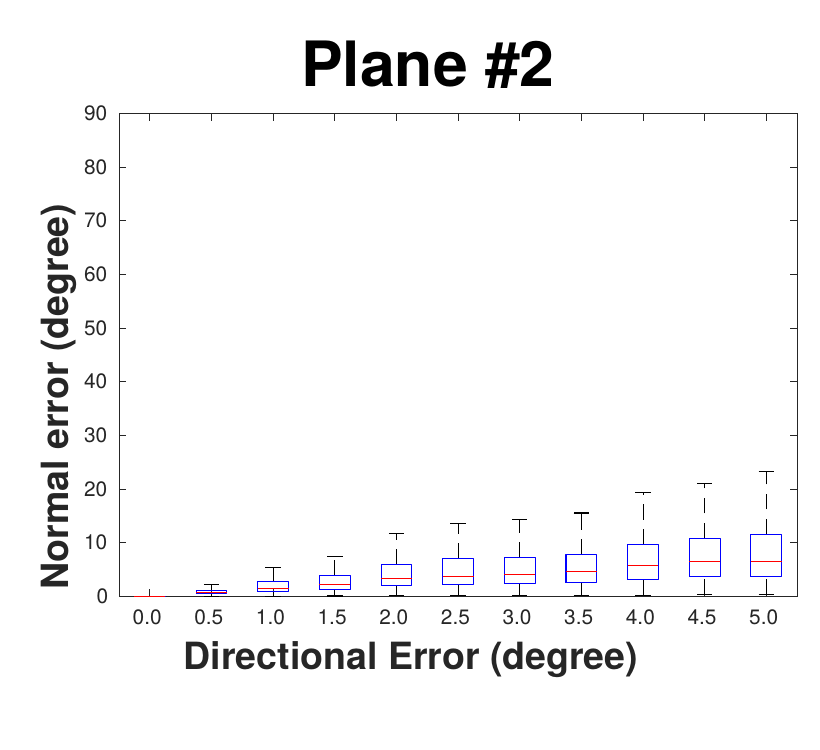}
\includegraphics[scale=.27]{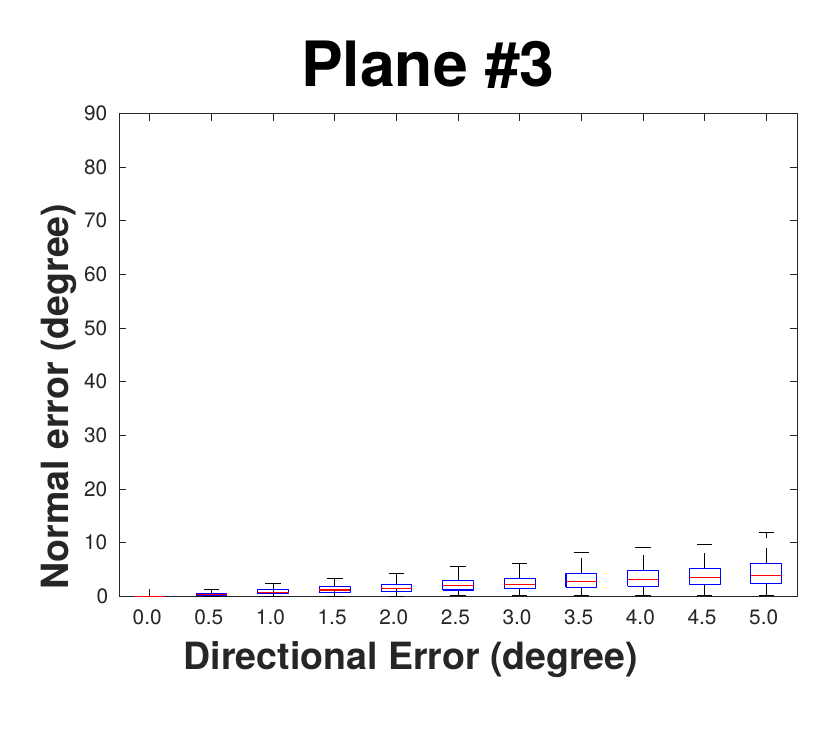}
  \caption{Standard Stereo}
  \label{fig:chart_synthetic_planar}
\end{subfigure}

\begin{subfigure}{1.0\textwidth}
  \centering
\includegraphics[scale=.27]{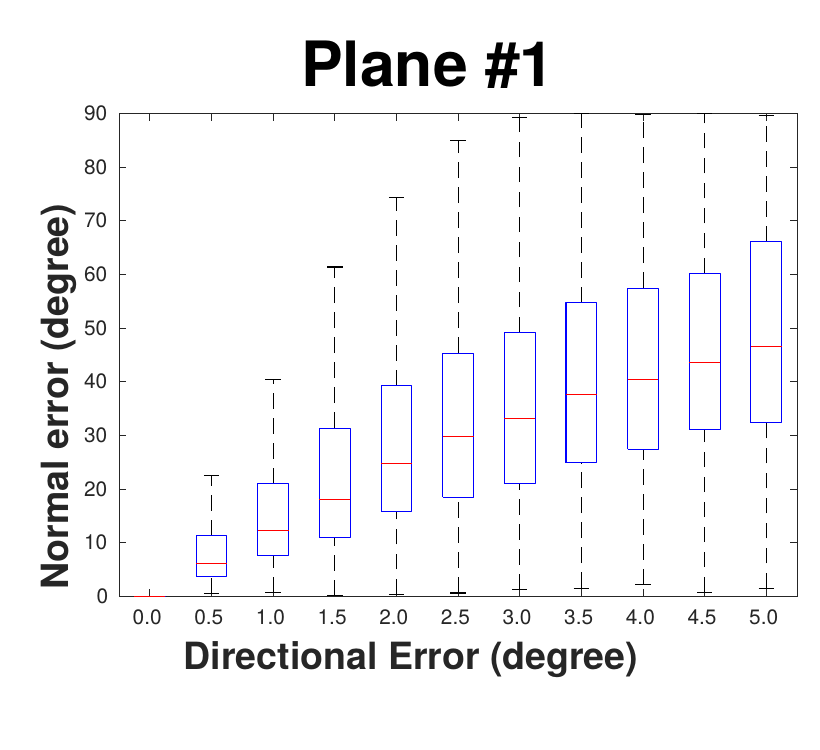}
\includegraphics[scale=.27]{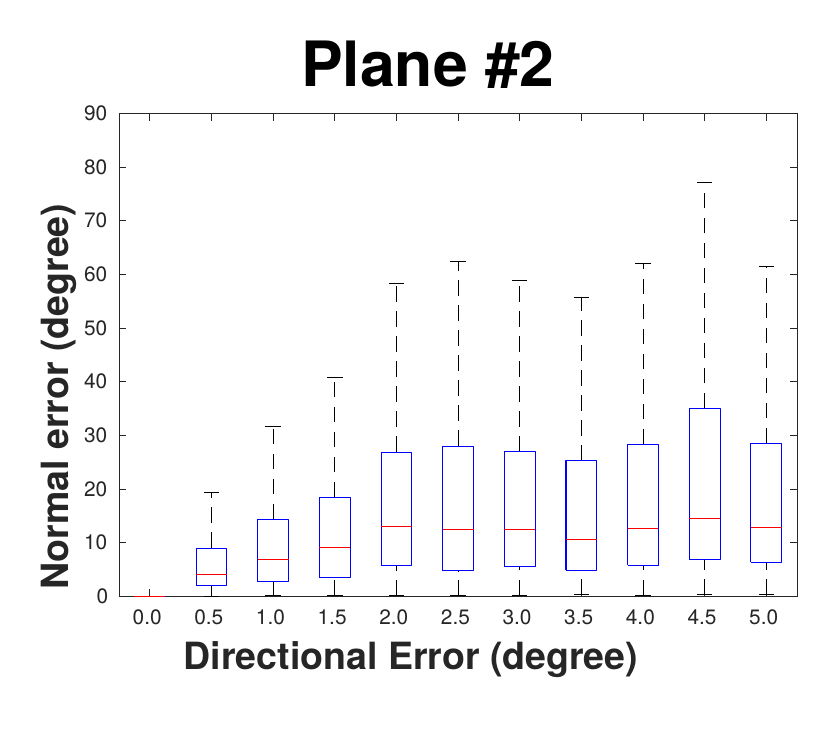}
\includegraphics[scale=.27]{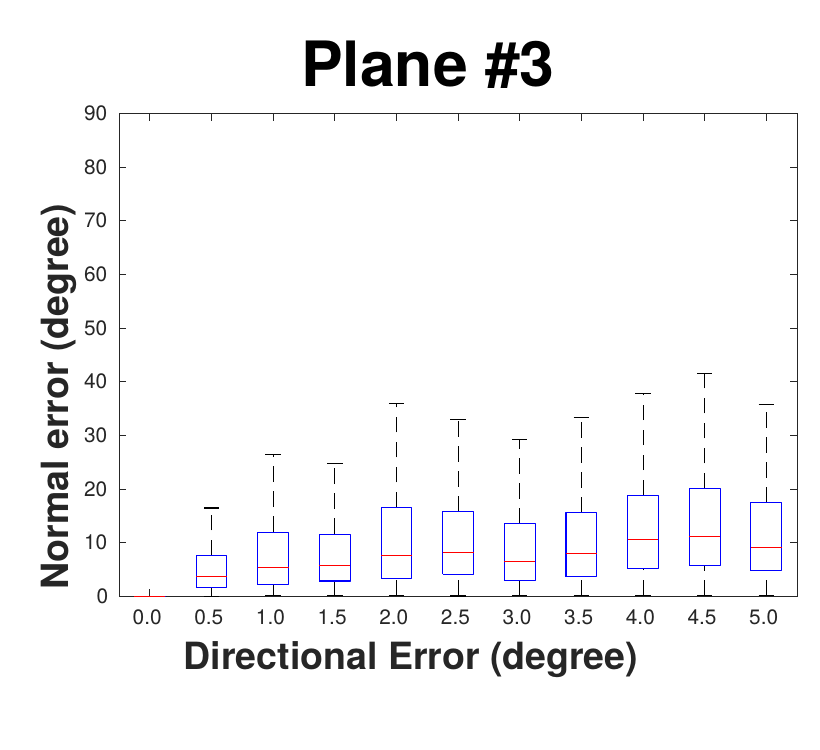}
  \caption{Forward Motion}
  \label{fig:chart_synthetic_planar}
\end{subfigure}
\caption{Synthetic testing results for special motion. Three different testing cases are considered: planar motion, forward motion and standard stereo.  The reconstruction error of tree planes are considered: \textbf{Plane \#1}: Parallel with camera image, \textbf{Plane \#2}: Horizontal plane, and \textbf{Plane \#3}: Vertical plane, parallel to the other ones}
\label{fig:chart_synthetic}
\end{figure*}

In Fig.~\ref{fig:chart_synthetic}, test results demonstrate that the accuracy highly depends on the orientation of the observed planes if special camera poses are considered. The horizontal dimension shows the directional noise level. The method F2UDIR is used to retrieve affine transformations at the sampled corner points. The numbering of the planes are visualized in Fig.~\ref{fig:synthetic_planes}: Plane \#1 is parallel to the first camera image, Plane \#2 is the horizontal one, and the vertical Plane \#3 is perpendicular to the other ones. The estimation of the normals for the first plane is very sensitive to the directional noise as it is seen in the left plots of Fig.~\ref{fig:chart_synthetic}. It can be also seen that there are differences in the camera poses: in case of forward motion, the accuracy is significantly lower than for the other types of stereo poses as Plane~\#2 and Plane~\#3 can be accurately estimated for both standard stereo and planar motion.

\subsection{Test on Real Images}
\begin{figure*}
     \centering
     \begin{subfigure}[b]{0.245\textwidth}
         \centering
         \includegraphics[width=\textwidth]{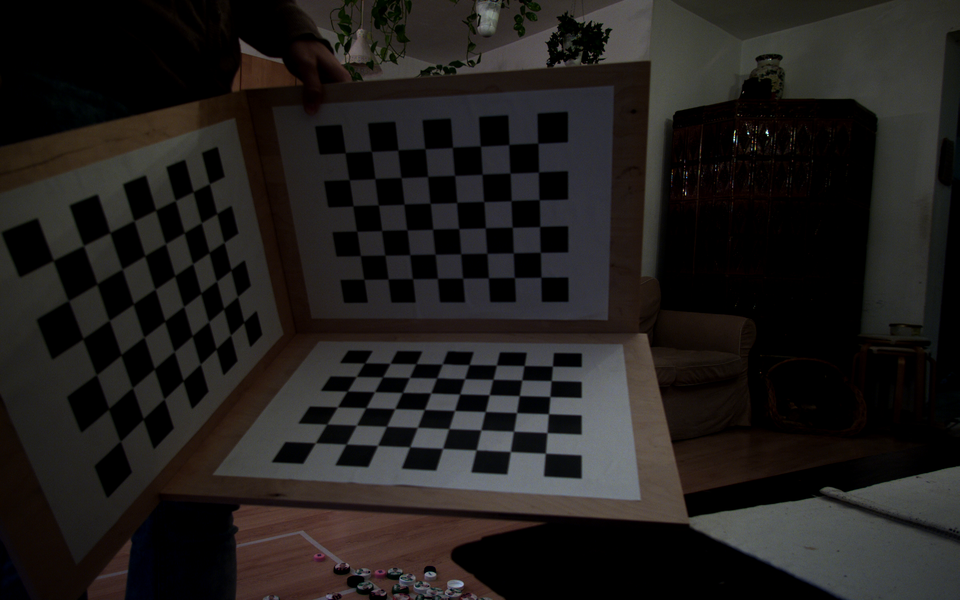}\vspace{0.6ex}
         \includegraphics[width=\textwidth]{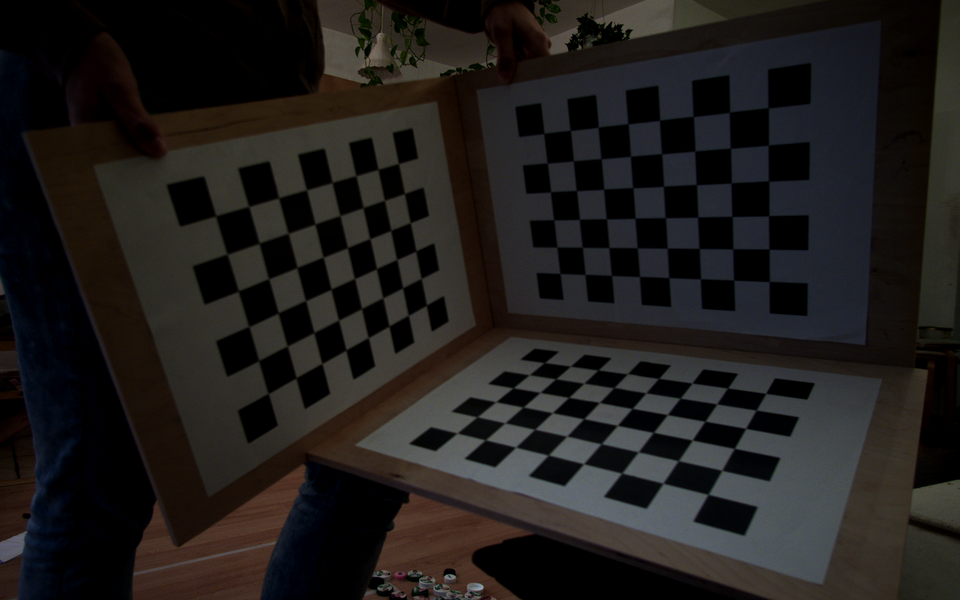}
         \includegraphics[width=\textwidth]{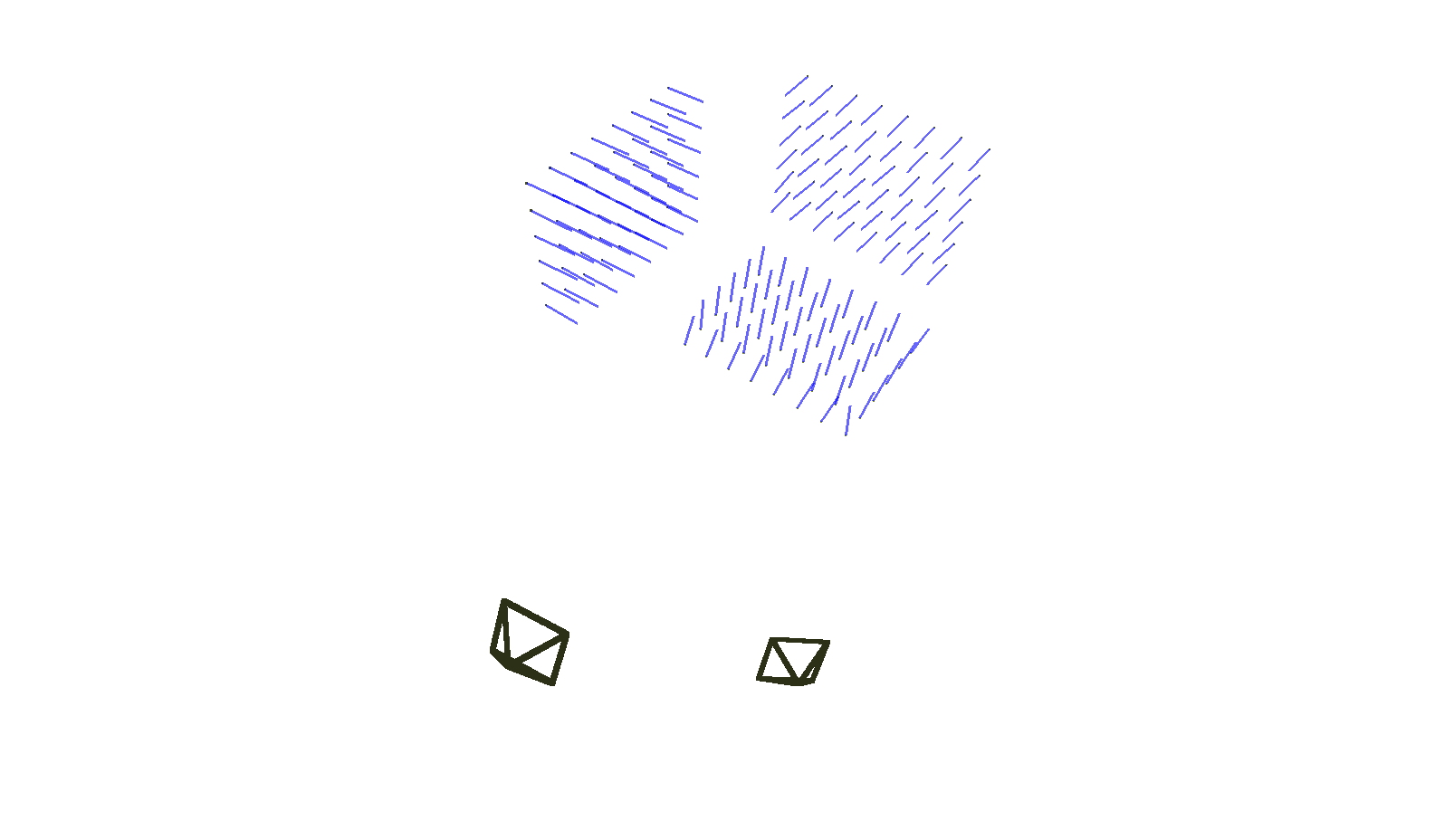}
         \caption{General}
         \label{fig:real_normal_general}
     \end{subfigure}
     \hfill
     \begin{subfigure}[b]{0.245\textwidth}
         \centering
         \includegraphics[width=\textwidth]{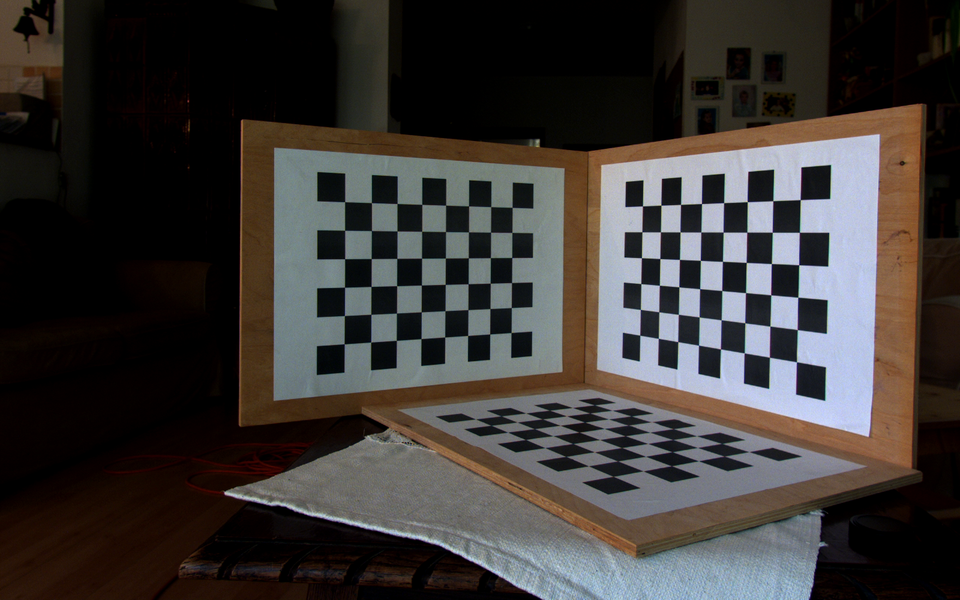}\vspace{0.6ex}
         \includegraphics[width=\textwidth]{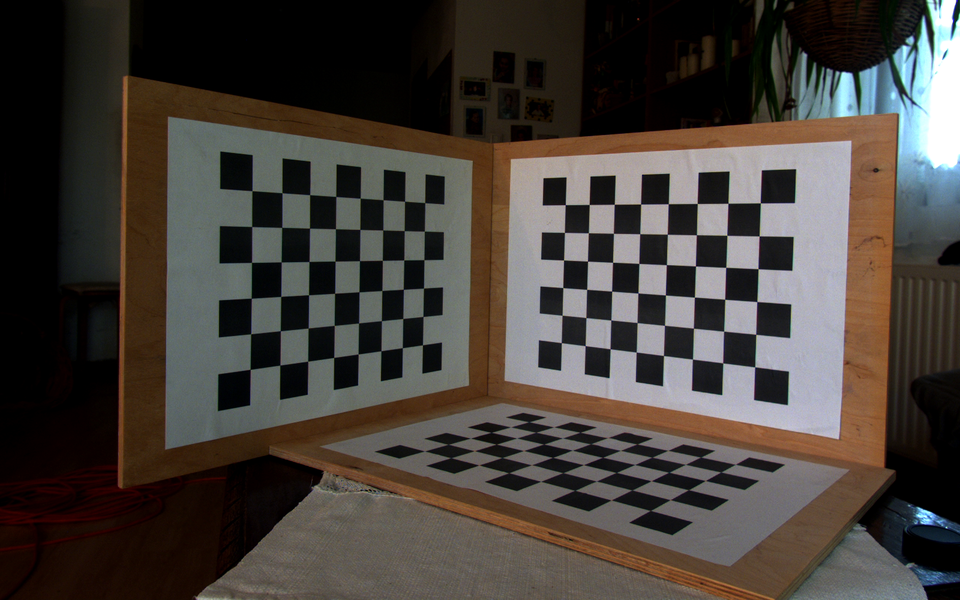}
         \includegraphics[width=\textwidth]{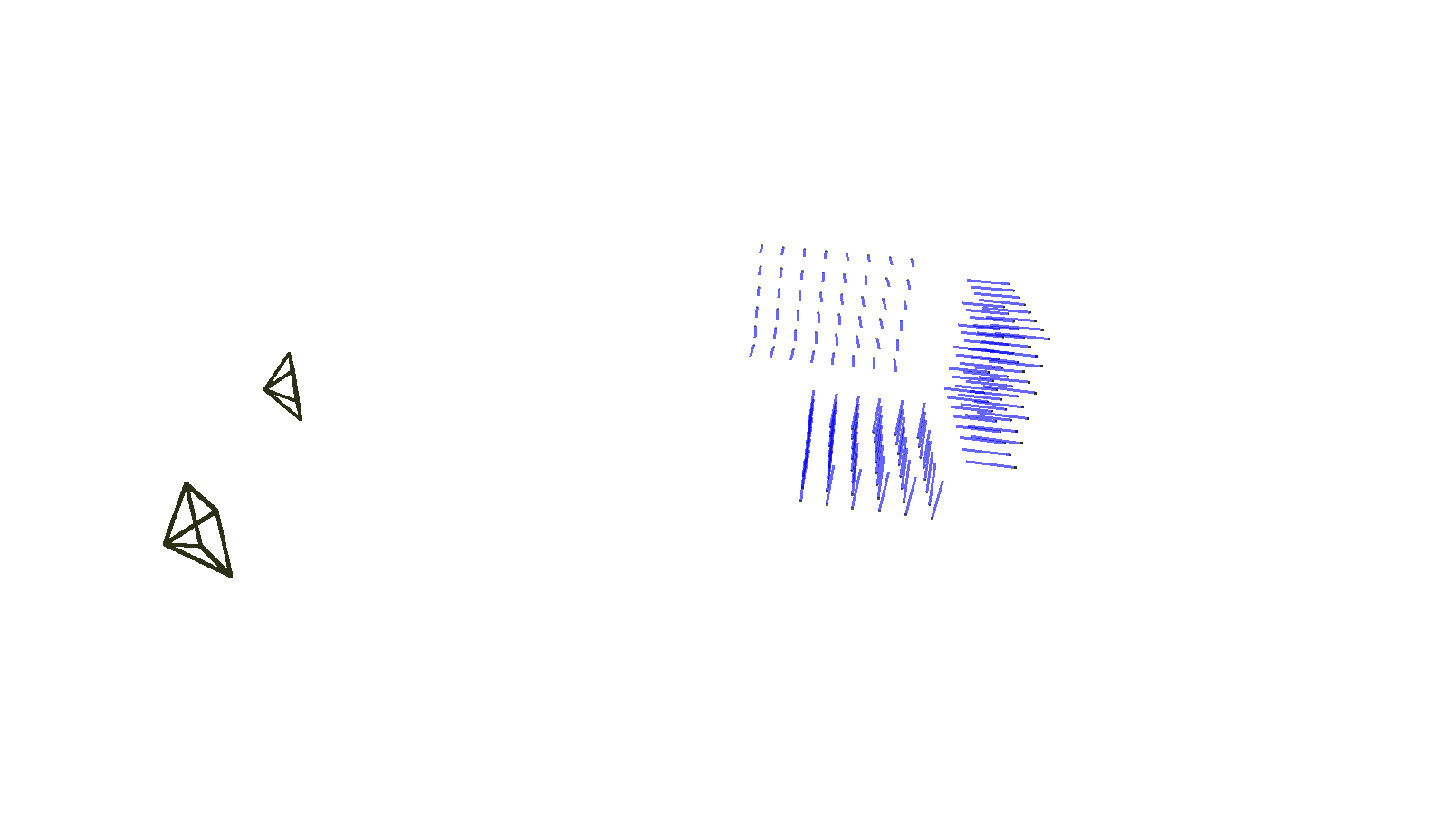}
         \caption{Planar motion}
         \label{fig:real_normal_planar}
     \end{subfigure}
     \hfill
     \begin{subfigure}[b]{0.245\textwidth}
         \centering
         \includegraphics[width=\textwidth]{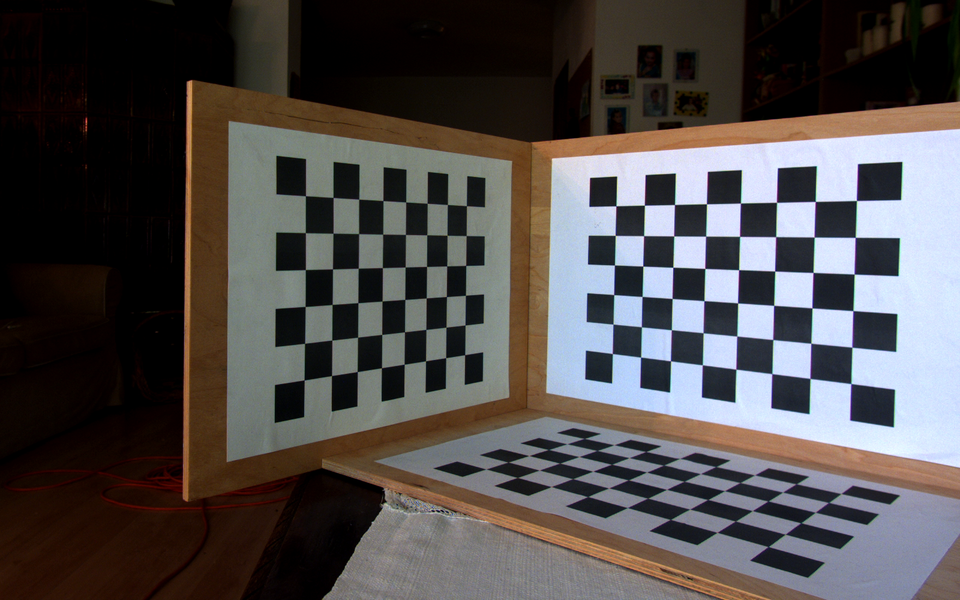}\vspace{0.6ex}
         \includegraphics[width=\textwidth]{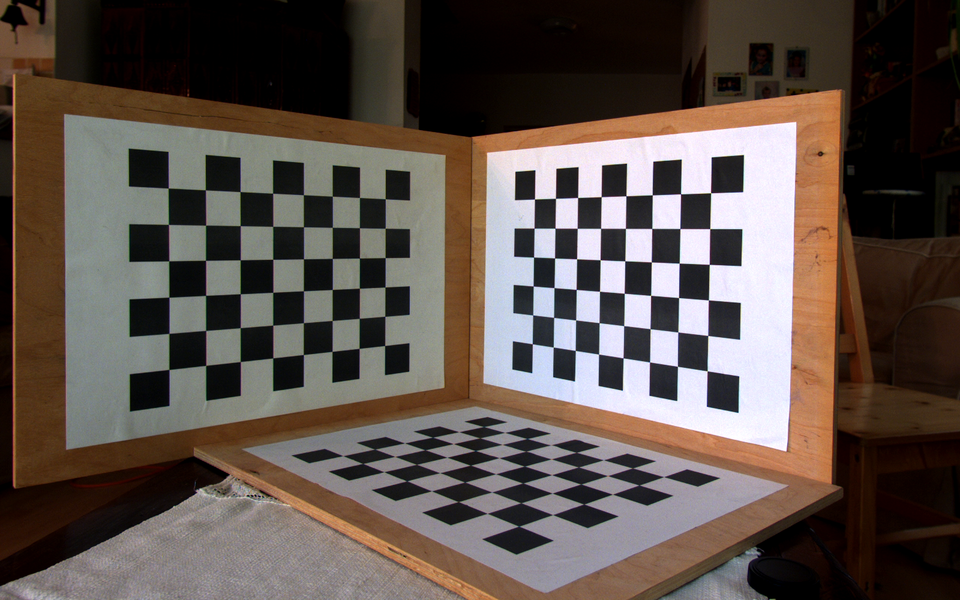}
         \includegraphics[width=\textwidth]{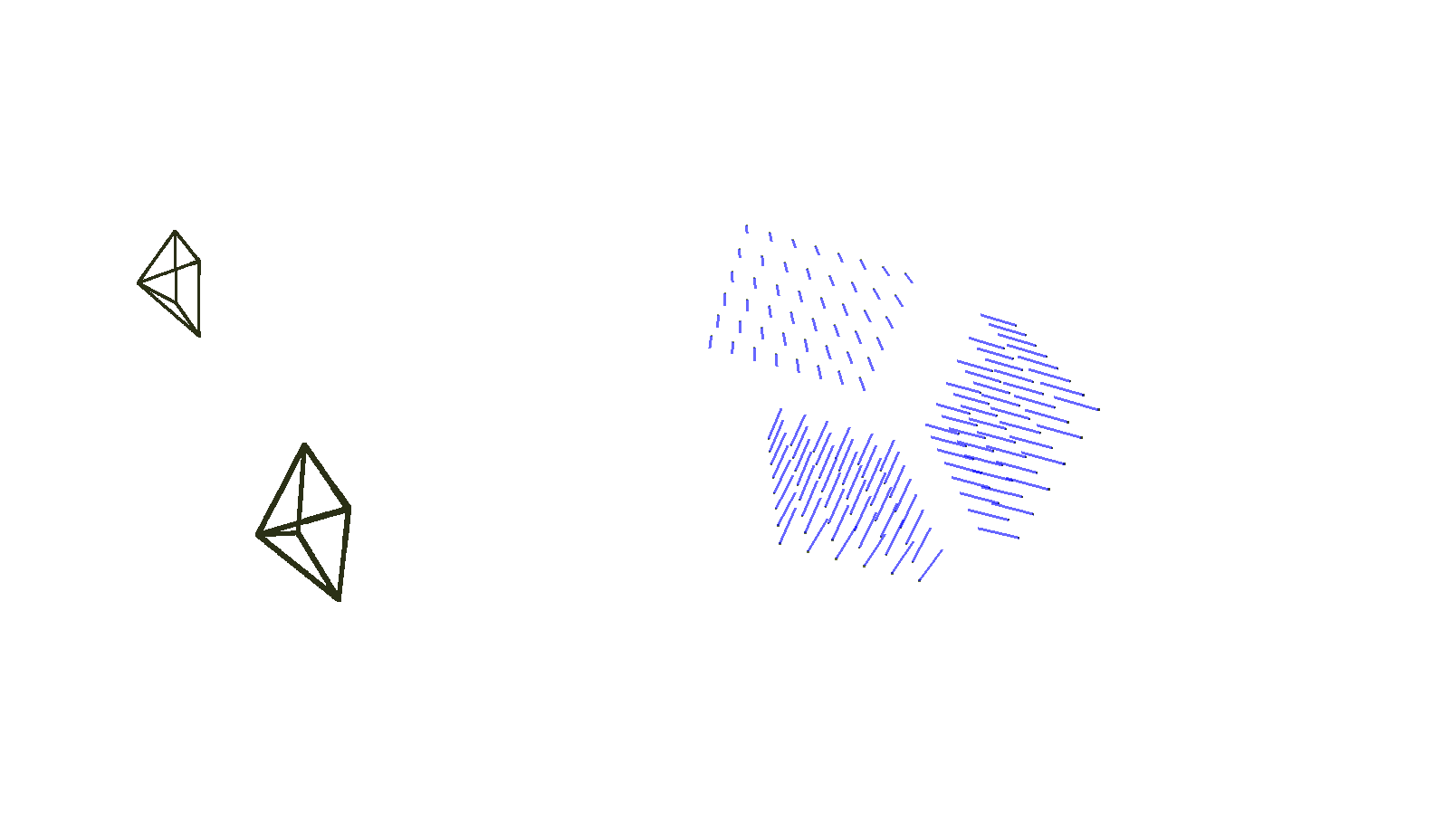}
         \caption{Standard stereo}
         \label{fig:real_normal_standard}
     \end{subfigure}
          \hfill
     \begin{subfigure}[b]{0.245\textwidth}
         \centering
         \includegraphics[width=\textwidth]{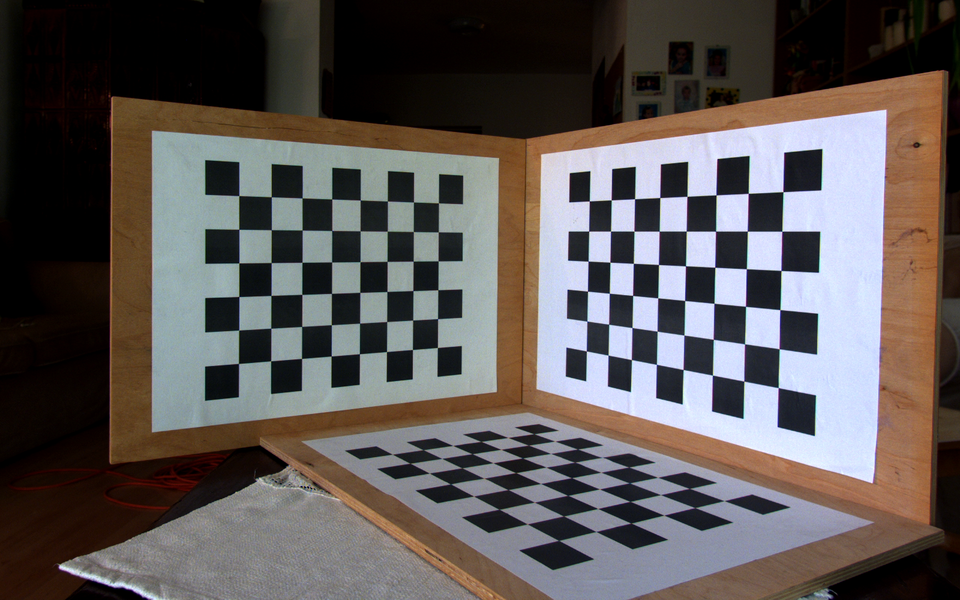}\vspace{0.6ex}
         \includegraphics[width=\textwidth]{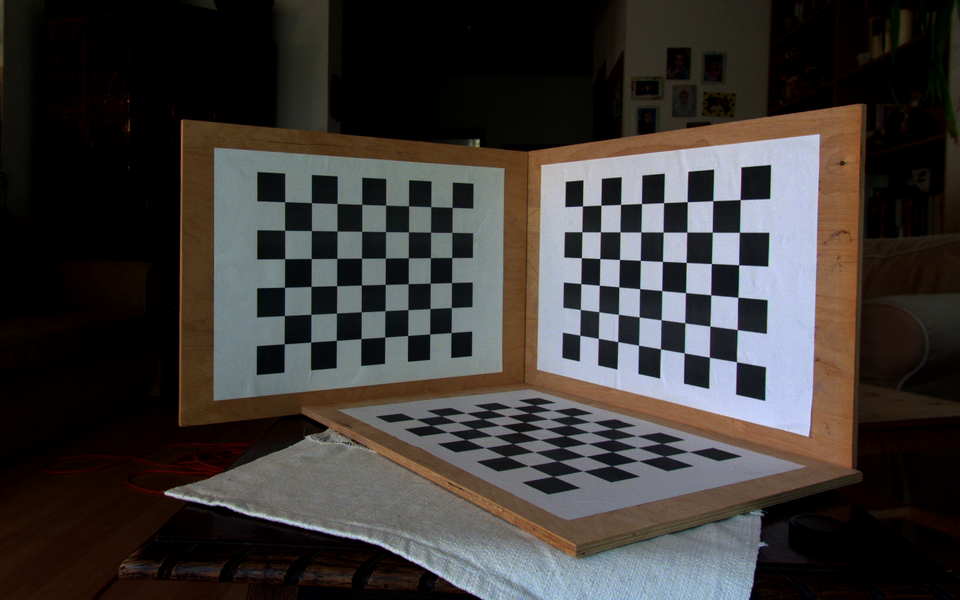}
         \includegraphics[width=\textwidth]{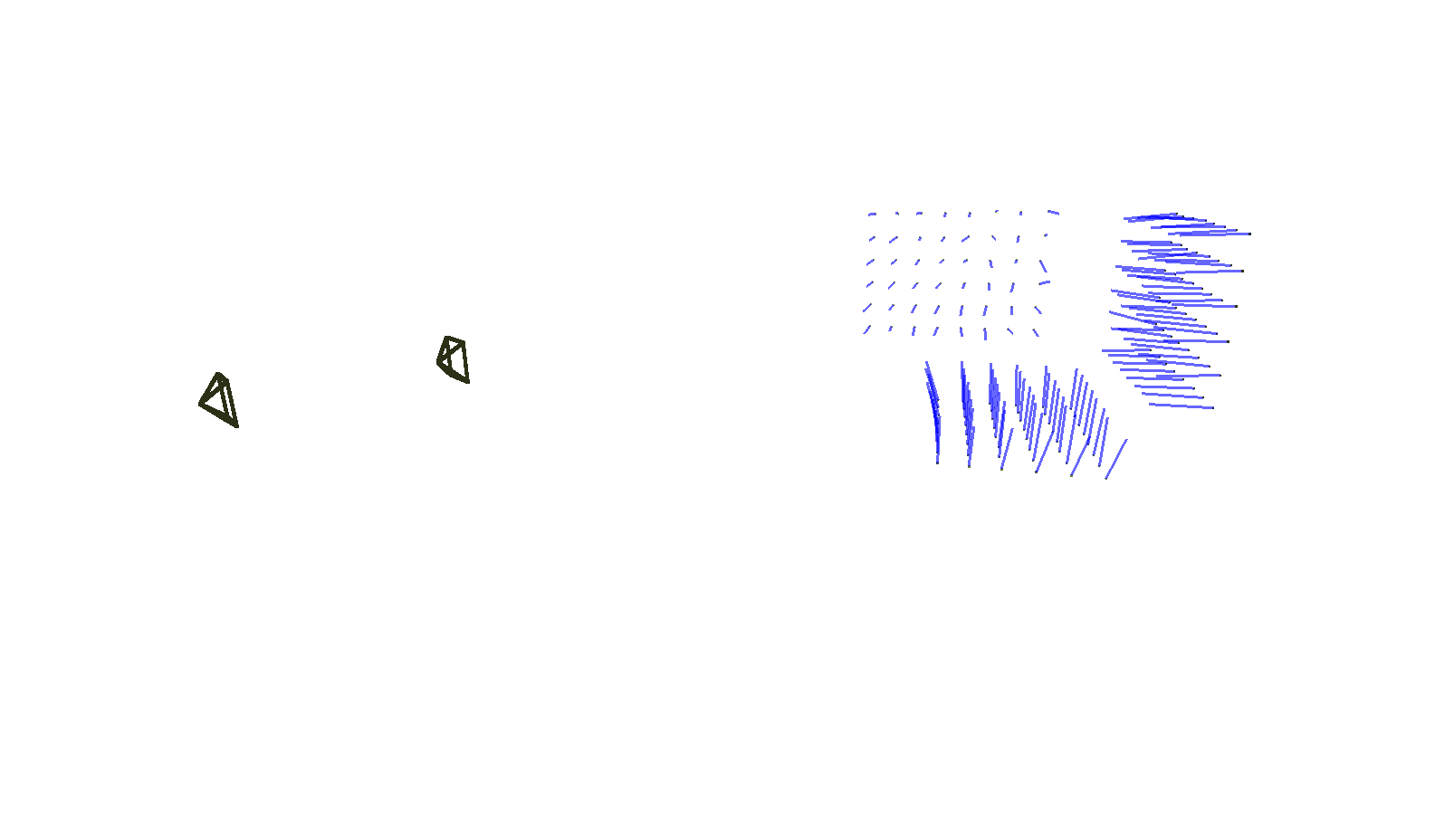}
         \caption{Forward Motion}
         \label{fig:real_normal_forward}
     \end{subfigure}
        \caption{Four test cases for quantitative comparison on real-world images. Regular optics are used. Top two rows: input images. Bottom row: a snapshot of the visualized 3D scene. Cameras are visualized by pyramids.}
        \label{fig:real_normal}
\end{figure*}

\begin{figure*}
     \centering
     \begin{subfigure}[b]{0.245\textwidth}
         \centering
         \includegraphics[width=\textwidth]{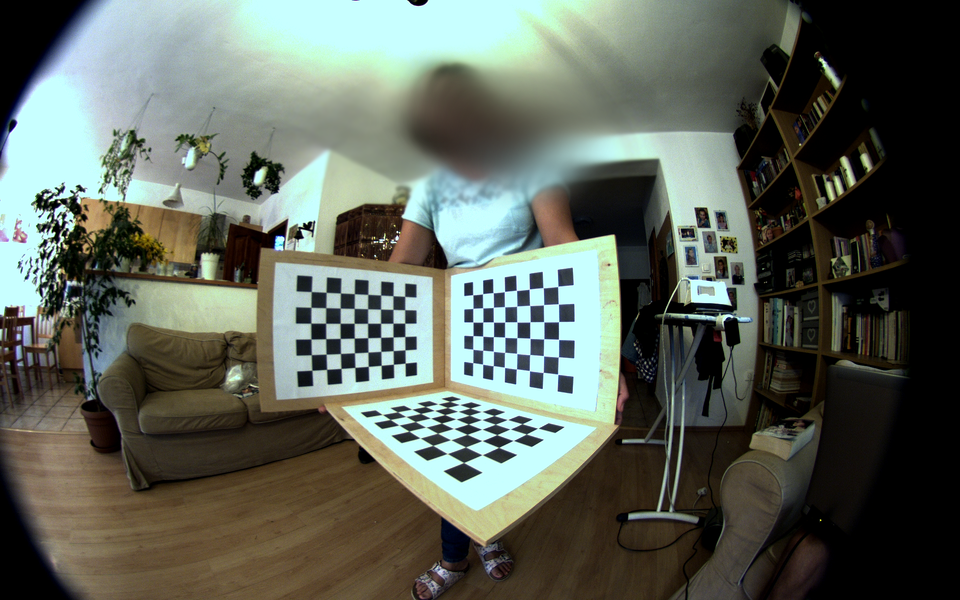}
         \includegraphics[width=\textwidth]{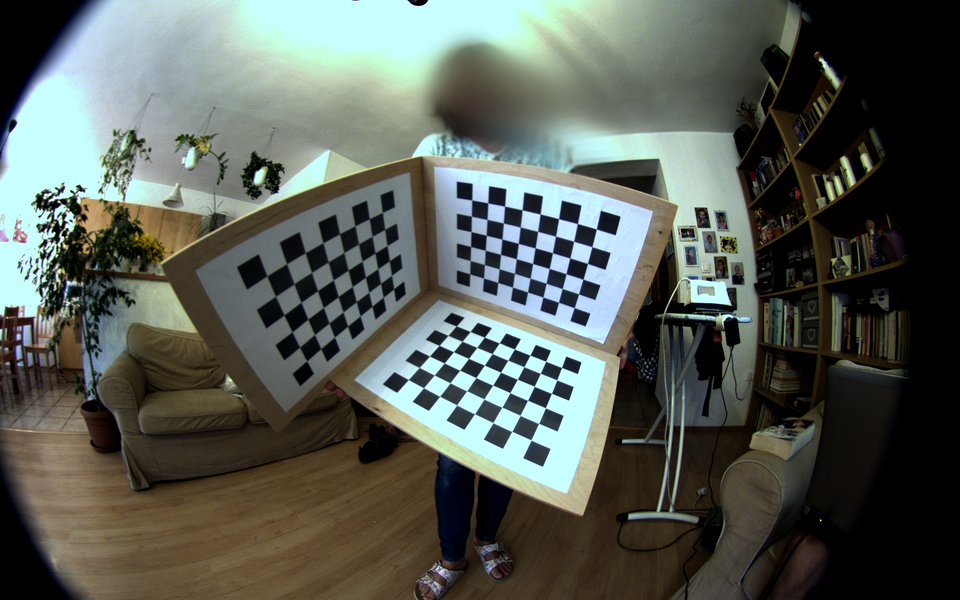}
         \includegraphics[width=\textwidth]{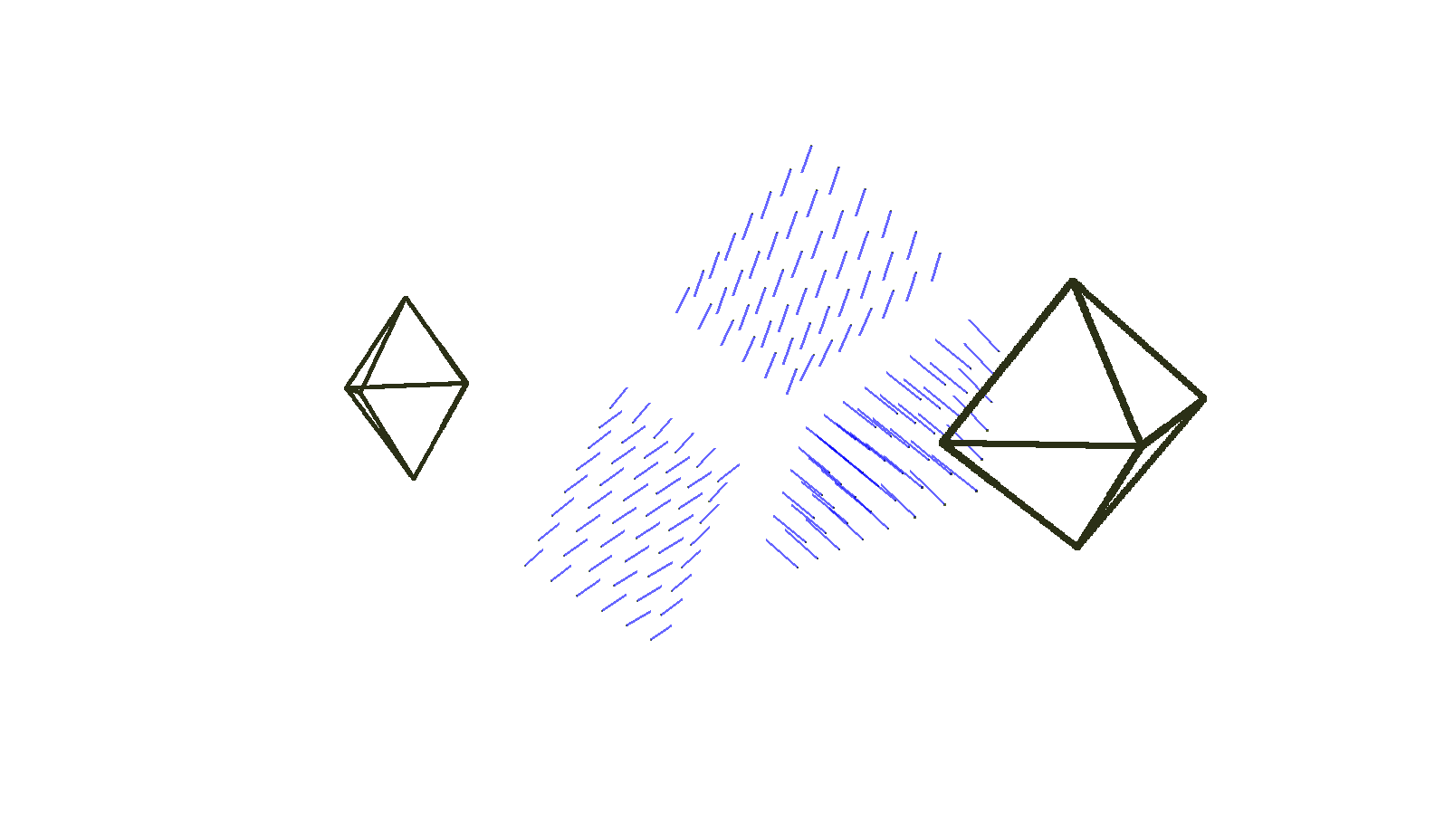}
         \caption{General}
         \label{fig:real_fisheye_general}
     \end{subfigure}
     \hfill
     \begin{subfigure}[b]{0.245\textwidth}
         \centering
         \includegraphics[width=\textwidth]{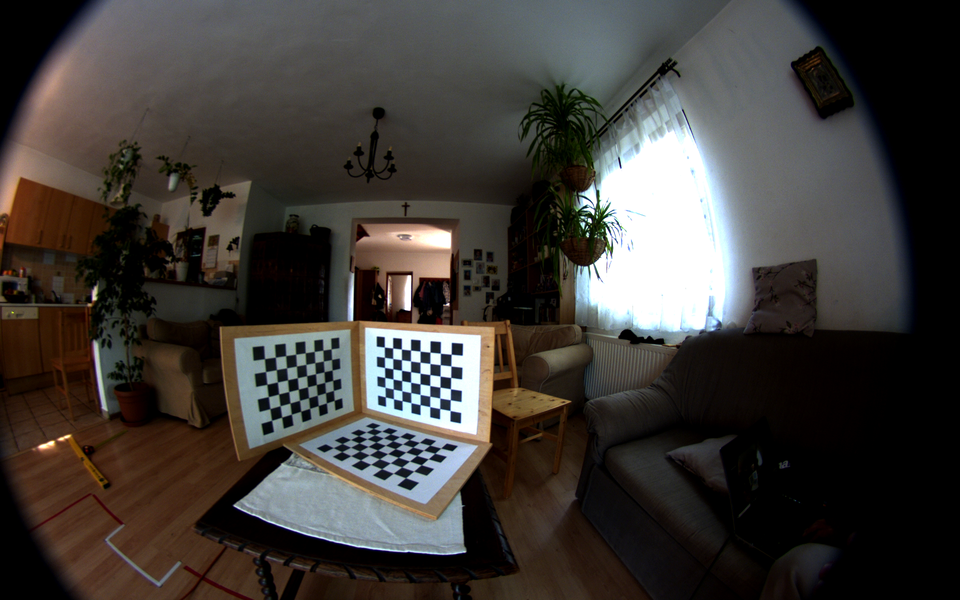}
         \includegraphics[width=\textwidth]{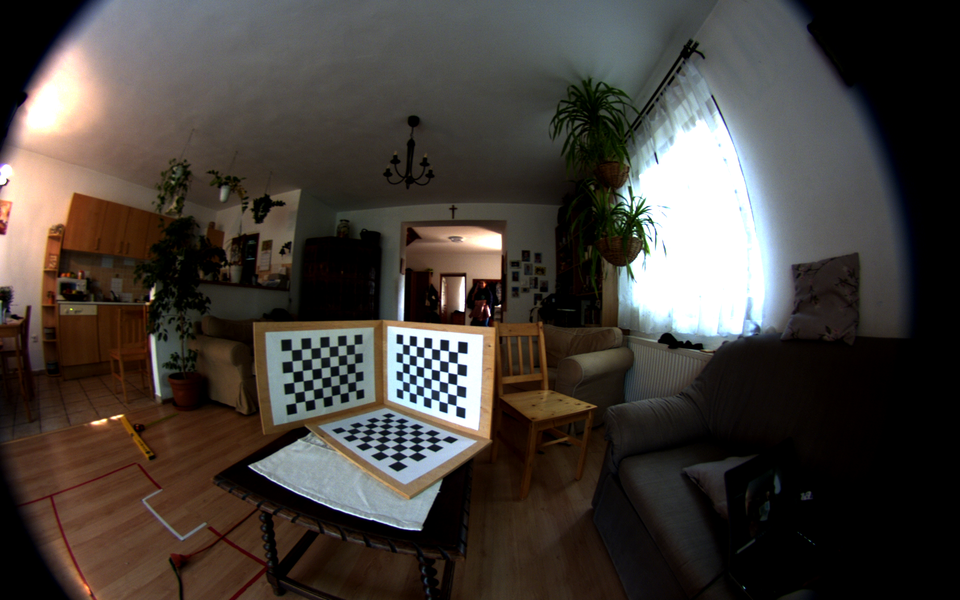}
         \includegraphics[width=\textwidth]{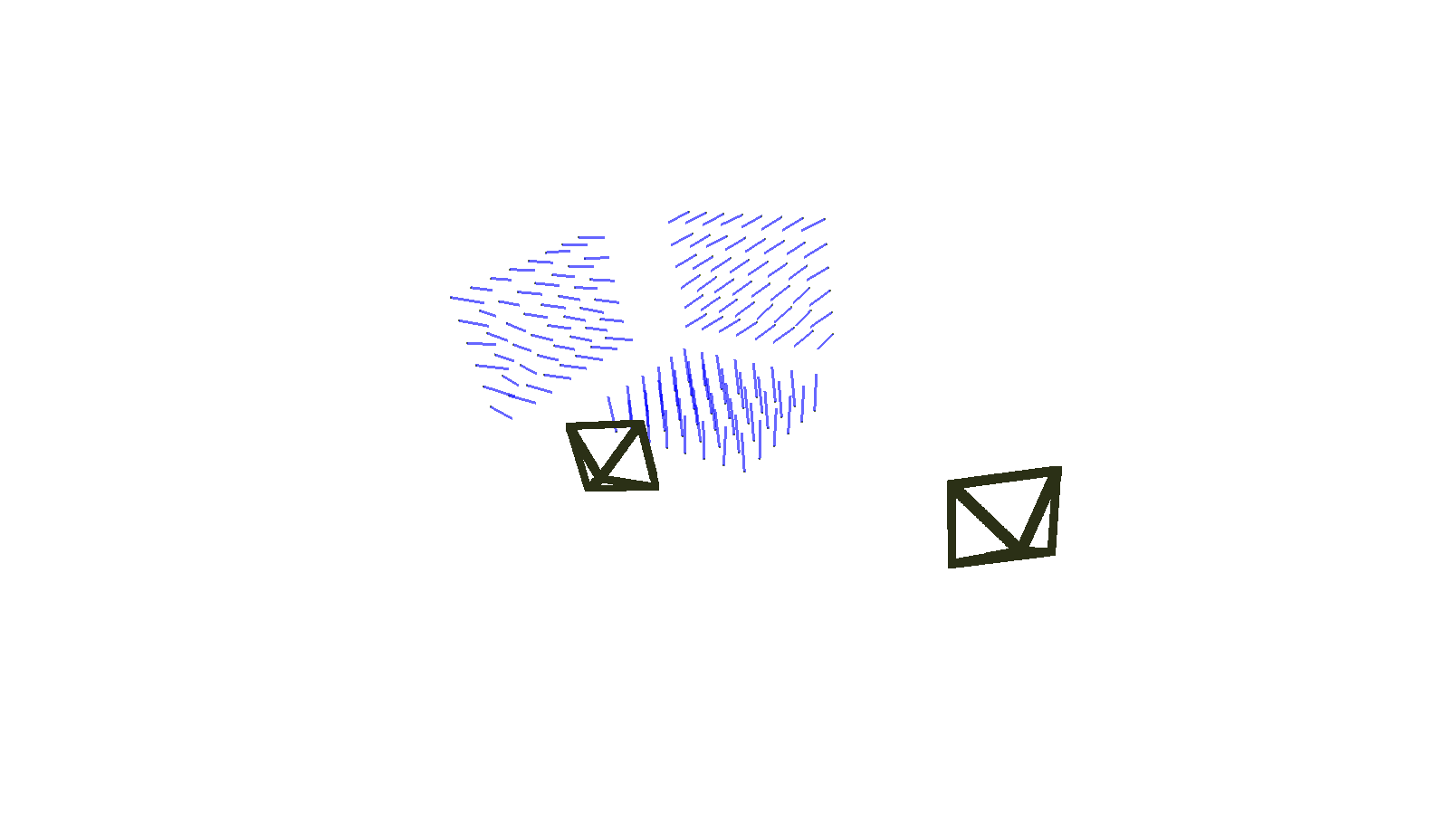}
         \caption{Planar motion}
         \label{fig:real_fisheye_planar}
     \end{subfigure}
     \hfill
     \begin{subfigure}[b]{0.245\textwidth}
         \centering
         \includegraphics[width=\textwidth]{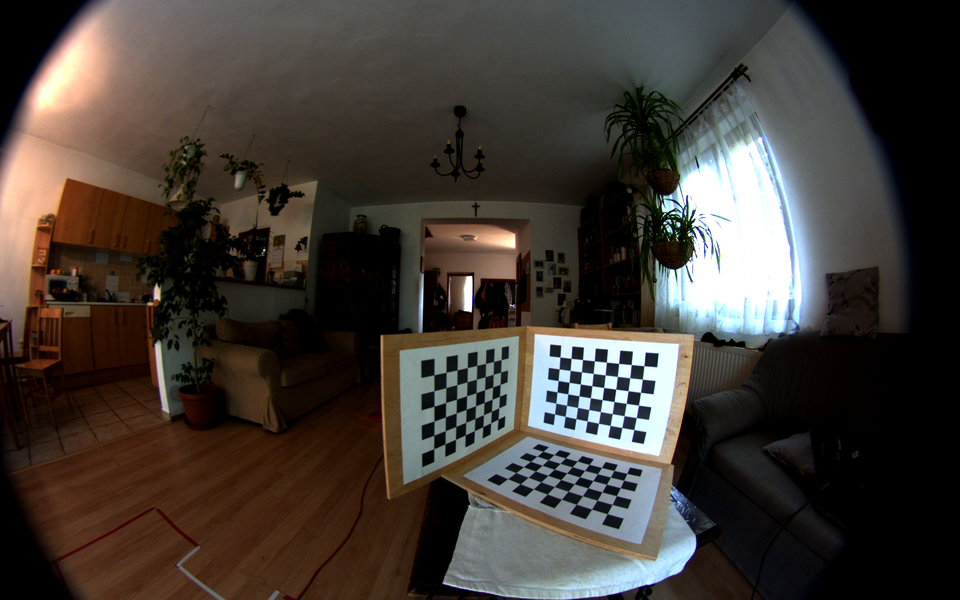}
         \includegraphics[width=\textwidth]{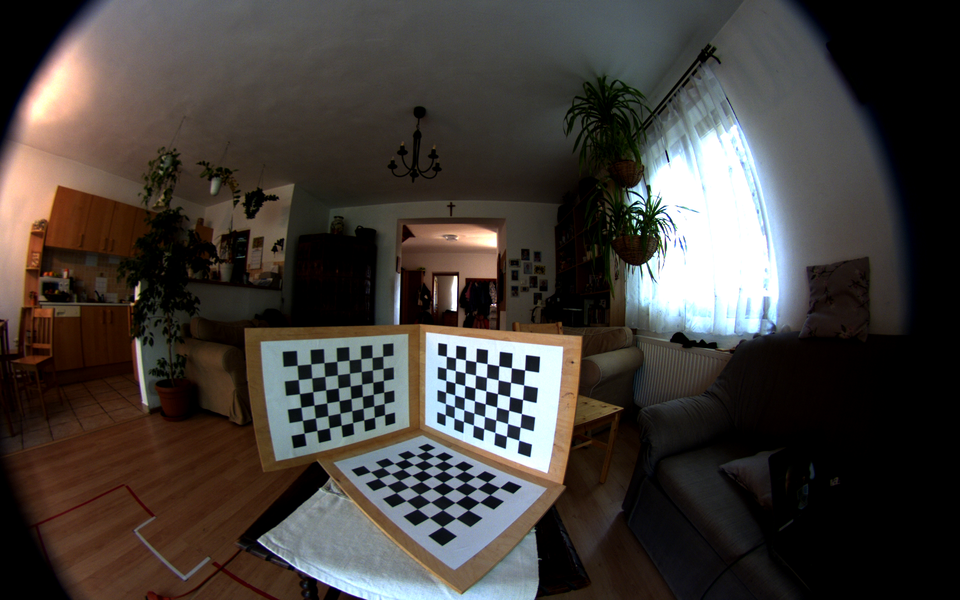}
         \includegraphics[width=\textwidth]{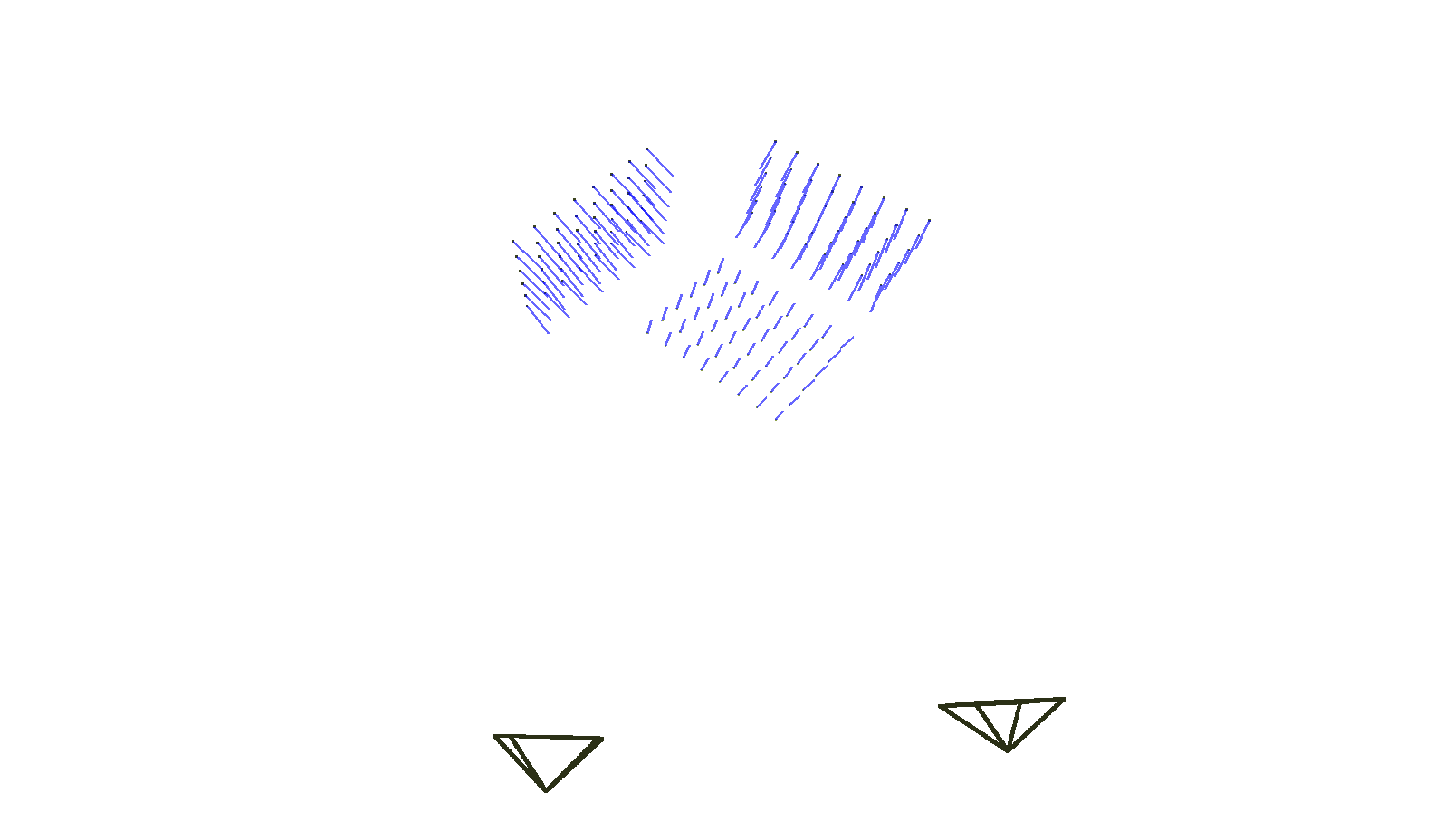}
         \caption{Standard stereo}
         \label{fig:real_fisheye_standard}
     \end{subfigure}
          \hfill
     \begin{subfigure}[b]{0.245\textwidth}
         \centering
         \includegraphics[width=\textwidth]{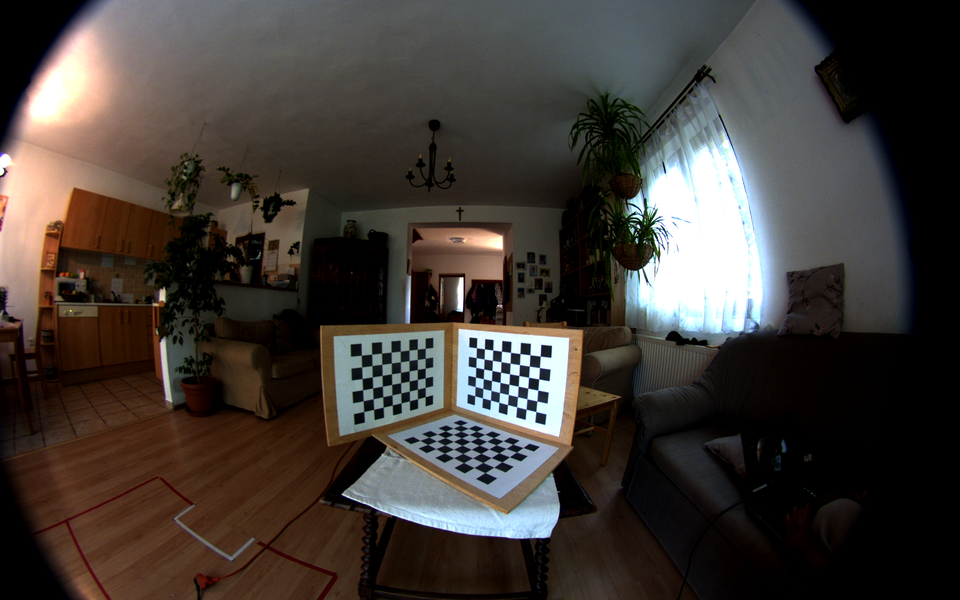}
         \includegraphics[width=\textwidth]{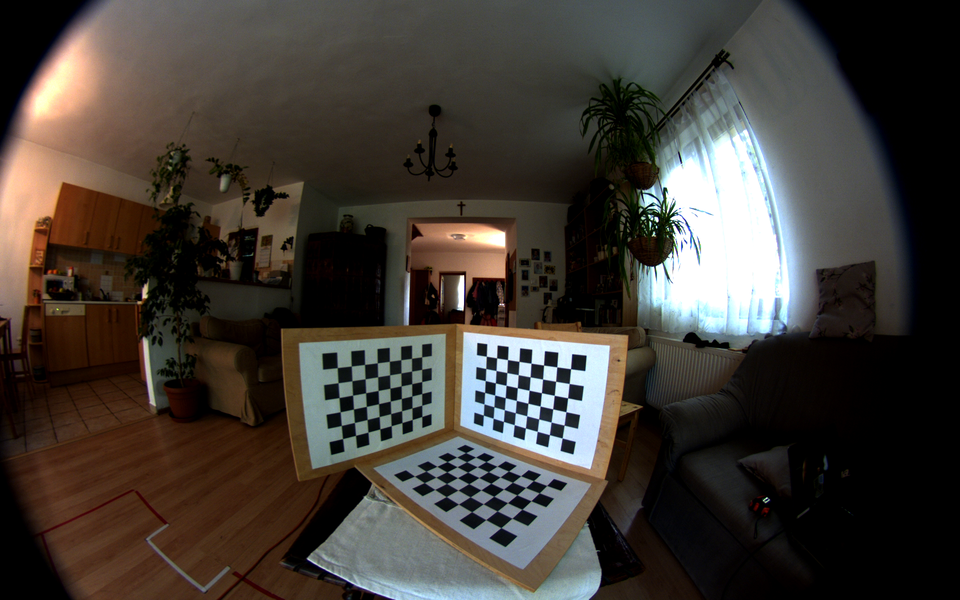}
         \includegraphics[width=\textwidth]{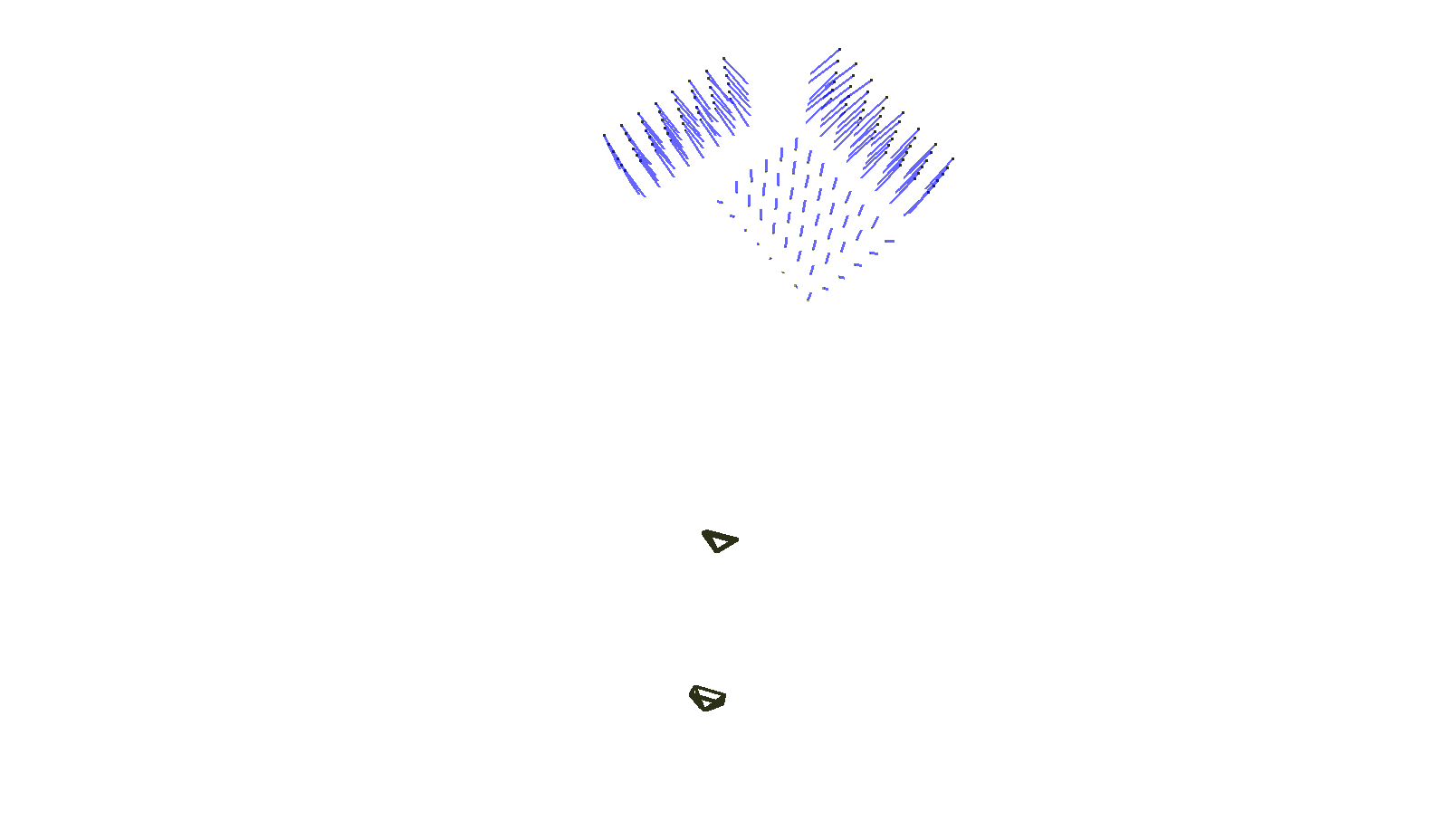}
         \caption{Forward Motion}
         \label{fig:real_fisheye_forward}
     \end{subfigure}
        \caption{Four test cases for quantitative comparison on real-world images. Fisheye lens is used with high field of view. Top two rows: input images. Bottom row: a snapshot of the visualized 3D scene. Cameras are visualized by pyramids.}
        \label{fig:real_fisheye}
\end{figure*}

To test how the lense properties affect the accuracy, a digital camera with two different types of lenses is applied. Similarly to the synthetic tests, the same four pose setups are considered. The input stereo images are visualized in Figs.~\ref{fig:real_normal} and~\ref{fig:real_fisheye}. We applied a HKVision MV-CA020-20GC camera with regular\footnote{Type:Fujicon SV-0614H, focal length:$6.1$ mm} and fisheye lens\footnote{Type: Fujicon FE185C086HA-1 , focal length:$2.7$ mm} for the tests. The same optics are used for the images of the same stereo setup.

\begin{figure}[ht]
\begin{subfigure}{.49\textwidth}
  \centering
\includegraphics[width=.5905\textwidth]{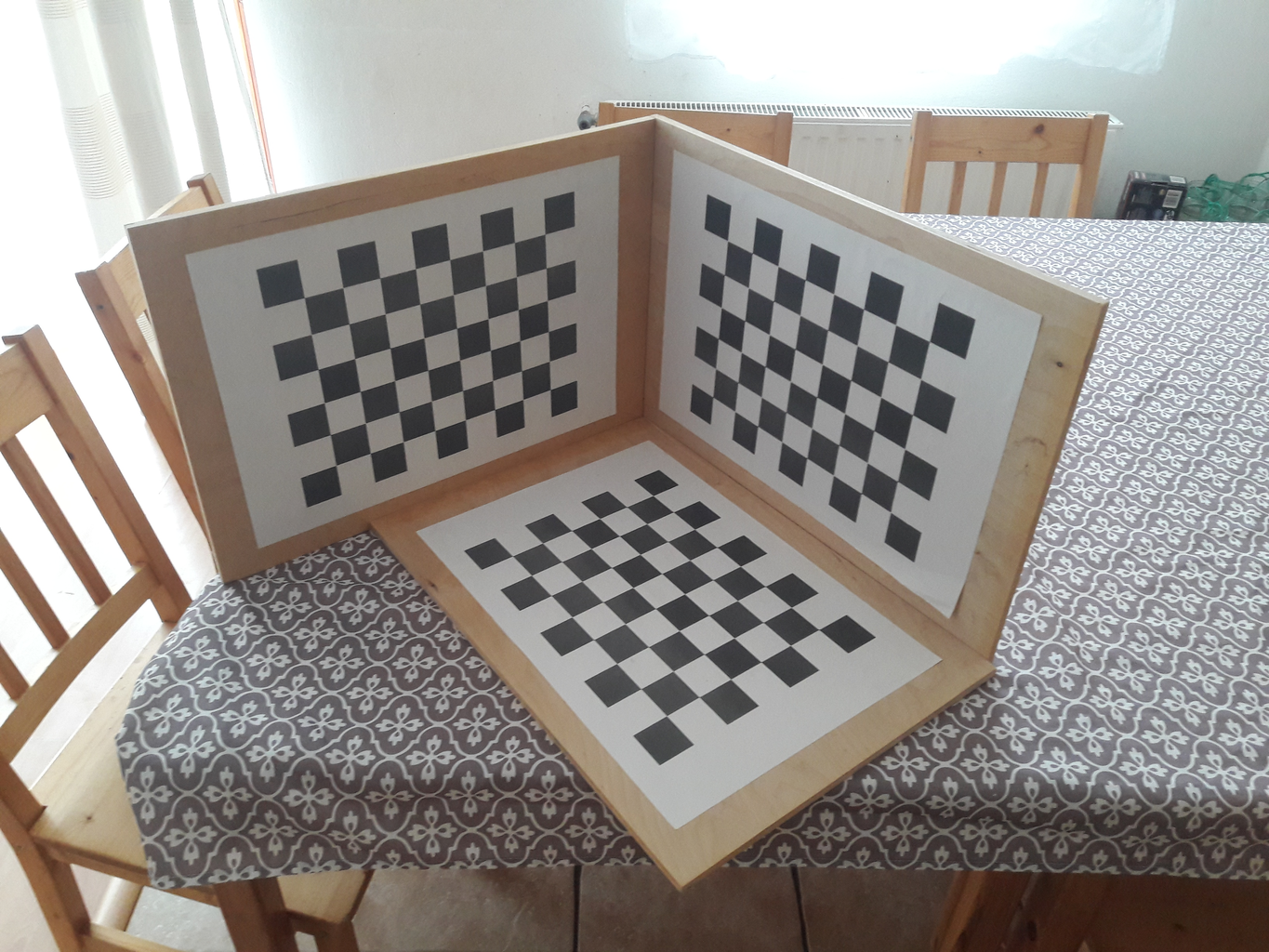}
  \caption{The testing 3D object is three perpendicular planes with chessboard patterns.}
\label{fig:directions}
\end{subfigure}
\begin{subfigure}{.49\textwidth}
  \centering
\includegraphics[width=.8\textwidth]{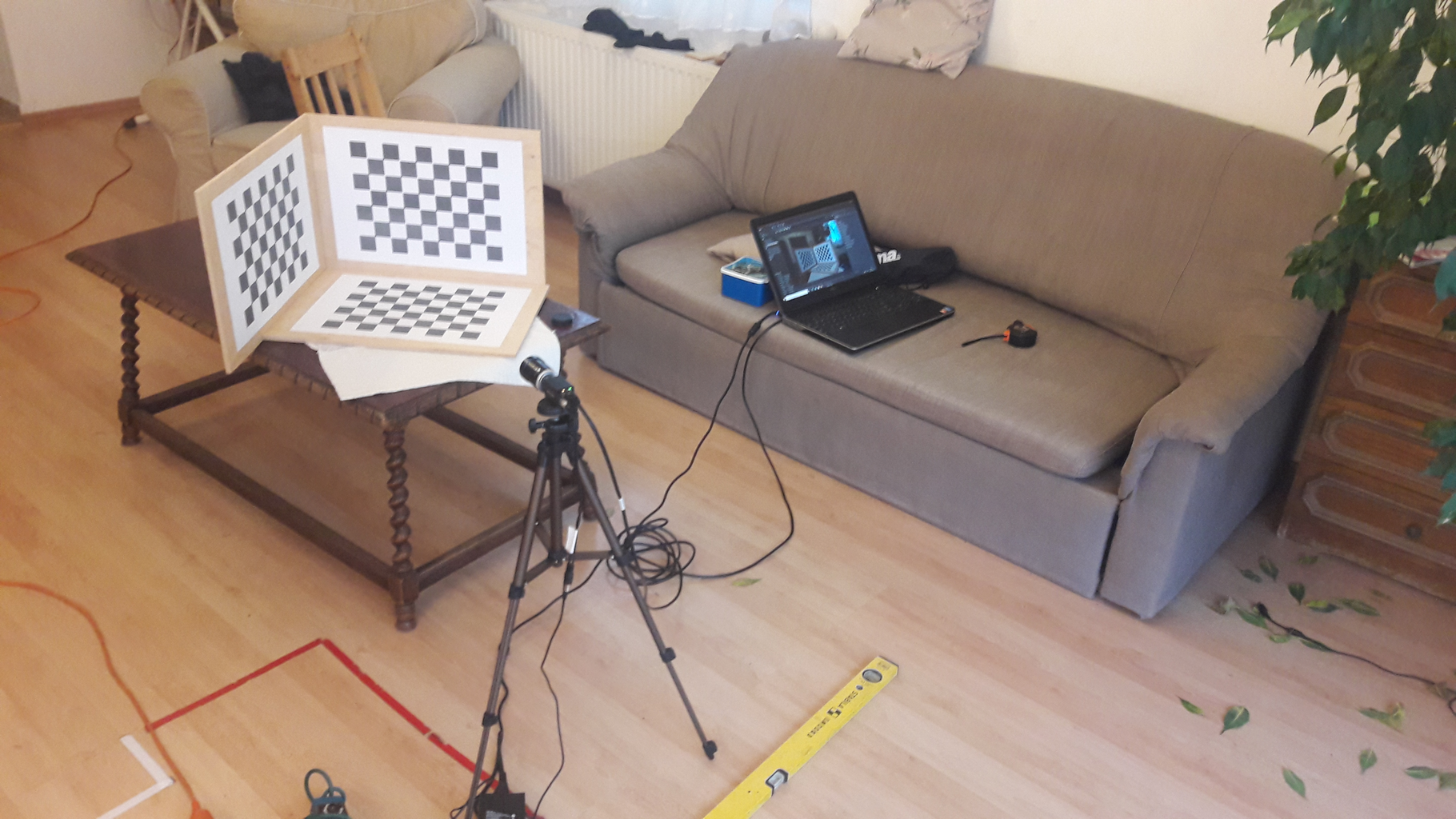}
\caption{The camera are mounted on a tripod. Camera orientation is measured by a spirit level (yellow).}
\label{fig:synthetic_planes}
\end{subfigure}
\caption{Testing environment for real-world experiments. Images are taken at 'home laboratory'.}
\label{fig:hom_lab}
\end{figure}

Due to the epidemic of the new coronavirus COVID-19, the images are taken at home. We call this special working place as 'home laboratory'. The testing environment is pictured in Fig.~\ref{fig:hom_lab}.

We calibrate the fisheye camera model with the frequently used method of \cite{DScaramuzzaMS06}. The 3D reconstruction is computed from the rectified images. We also tried the built-in calibration technique from OpenCV\footnote{http://www.opencv.org} using radial and tangential distortion parameters, however, OpenCV's method is not accurate enough as non-perspective distortion parameters are calculated only in the image space contrary to the algorithm of \cite{DScaramuzzaMS06} where a more realistic 3D model is applied for considering non-perspective distortion.

First, a validation process is carried out. As the real size of the chessboard fields is known, metric reconstruction is also possible. The baseline of a stereo setup can be measured in the standard and forward motion cases: only the locations of the tripod\footnote{The cameras are attached to a tripod.} should be measured. Then the reconstructed baseline length can be compared to the ground truth values. All the five variants of affine estimation, proposed in Section~\ref{sec:estimation}, are tested. The resulting baseline lengths are reported in Table~\ref{tab:real_distances}. The precision of the real measurement is a few millimeters, therefore, we think that all the resulting qualities are suitable for most of the applications. There are not significant differences in terms of baseline accuracy, as the baseline lengths are principally influenced by the PCs. Remark that the baseline error is significantly larger when OpenCV's calibration method is used for fisheye lens, especially, when the cameras undergo forward motion.

As it is known that the scene consists of three perpendicular planes, quantitative comparison is also possible for real data. The results are reported in Table~\ref{tab:real_main_results}. For real-world tests, Plane \#2 is the horizontal one, Plane \#1 and \#3 are on the left and right sides, respectively. 

All the proposed stereo poses are realized for the real tests. For the forward, planar and standard cases, it is very important that the image planes must be perpendicular to the ground as it is required by the fundamental matrix estimation technique we use \cite{Ortin2001}. The perpendicularity w.r.t. ground is set by a standard spirit level. However, it is found that this orthogonality is not perfect for the planar motion with fisheye lens; if the estimation of a general fundamental matrix is applied, more accurate results are obtained. We include the most accurate results in the table for this and only this case.

The comparison is the same as for the synthetic tests: the chessboard planes are estimated from the points by PCA, the normals are compared to the corresponding plane normals, the error is written in degrees. Our first impression about the real tests is that the results are very convincing for the camera with normal optics: the perpendicularity error of the planes are below $10^{\circ}$, the maximum normal error is also only a few degrees. It is very interesting that the standard stereo is the test case for which the most accurate normals are obtained, while the forward motion is the test case with less accurate normal estimation. However, the same conclusion was drawn for the synthetic tests, as it is discussed in the previous section.

For real tests, none of the vertical planes are not parallel to the image planes. However, Plane \#2 is close to be parallel, and the accuracy for normal estimation is lower than that for other planes. This statement is true for both forward and planar motions. Thus, in conjunction with the synthetic tests, it can be stated that the quasi-parallel planes cannot be reconstructed accurately.

For the real tests, the results can also be evaluated qualitatively. The reconstructed 3D scenes are visualized in the bottom of Figs.~\ref{fig:real_normal} and~\ref{fig:real_fisheye}. The results are of high quality, the perpendicular planes are successfully reconstructed, significant distortions are not seen. It is also clearly seen that the reconstructed normals are not perfect for the special poses as it is discussed above.


\begin{table}[ht]
\caption{Quantitative results for real images. Four stereo pose types are tested. Both regular (Reg.) and fisheye (Fish.) lens are applied. \textit{Orthogonality error}: The captured planes are perpendicular, the deviation from $90^{\circ}°$ is reported. \textit{Normal Dir. Err.} The estimated normals compared to the normal of the planes fitted by PCA algorithm. Angular errors are reported in degrees.}
\begin{center}
\begin{tabular}{c c c|c c|c}
\multicolumn{2}{c}{\multirow{2}{*}{~}}& \multicolumn{2}{c}{\textbf{ General Motion}} & \multicolumn{2}{c}{\textbf{Planar Motion}}  \\
\multicolumn{2}{c}{}& Reg. & Fish. & Reg. & Fish. \\
\hline
    \multicolumn{1}{c}{\multirow{3}{*}{\textbf{Orthogonality Error}}}&Plane \#1--\#2 & 3.6010   & 0.7112     &  4.7177  & 0.5222    \\
    \multicolumn{1}{c}{}&Plane \#1--\#3&  0.2599   & 0.2068    & 2.1562 &   2.0352     \\  
    \multicolumn{1}{c}{}&Plane \#2--\#3&  4.1743 &  1.1883 & 3.1345 &  2.0037   \\
    \hline
    \multicolumn{1}{c}{\multirow{3}{*}{\textbf{Normal Dir. Err. (mean)}}}&Plane \#1 & 0.8622   & 2.8618   &   0.4945      & 4.7222   \\
    \multicolumn{1}{c}{}&Plane \#2&  5.1772    &  0.9905   &    4.1530   &  8.1205     \\  
    \multicolumn{1}{c}{}&Plane \#3& 3.6114 &  0.6055 &    1.2787 &  4.1878 \\    
    \hline
    \multicolumn{1}{c}{\multirow{3}{*}{\textbf{Normal Dir. Err. (median)}}}&Plane \#1 &  0.6770    &  1.7505    &  0.4129    & 4.9323     \\
    \multicolumn{1}{c}{}&Plane \#2&  5.0156    &  0.9453    &    3.8251    & 8.0500 \\  
    \multicolumn{1}{c}{}&Plane \#3& 3.7265 & 0.4920 & 1.1673 & 3.7898  \\    
    \hline
\end{tabular}
\end{center}

\begin{center}
\begin{tabular}{c c c|c c|c}
\multicolumn{2}{c}{\multirow{2}{*}{~}} & \multicolumn{2}{c}{\textbf{Standard Stereo}} & \multicolumn{2}{c}{\textbf{Forward Motion}}  \\
\multicolumn{2}{c}{}& Reg. & Fish. & Reg. & Fish.   \\
\hline
    \multicolumn{1}{c}{\multirow{3}{*}{\textbf{Orthogonality Error}}}&Plane \#1--\#2 &  0.5946   & 0.6358     & 8.8721   & 6.9179  \\
    \multicolumn{1}{c}{}&Plane \#1--\#3&   1.1304   & 3.1221    &  5.3309    &   3.5328  \\  
    \multicolumn{1}{c}{}&Plane \#2--\#3&   0.6968 &  0.8731 & 3.5293 &   0.7534\\
    \hline
    \multicolumn{1}{c}{\multirow{3}{*}{\textbf{Normal Dir. Err. (mean)}}}&Plane \#1 &  0.3824    & 1.2891    & 6.6203    & 7.3735  \\
    \multicolumn{1}{c}{}&Plane \#2&   0.3802   &  1.9127  & 8.3326   &  7.4163  \\  
    \multicolumn{1}{c}{}&Plane \#3&   0.2025 &   0.9249 &  1.0241 &  1.334\\    
    \hline
    \multicolumn{1}{c}{\multirow{3}{*}{\textbf{Normal Dir. Err. (median)}}}&Plane \#1 &   0.3902   & 1.1158    &  6.0689  & 6.2085 \\
    \multicolumn{1}{c}{}&Plane \#2&   0.3724   & 1.6271    &   8.4207    &   7.5404  \\  
    \multicolumn{1}{c}{}&Plane \#3&   0.2013 & 0.7855 & 0.9124 &  1.2563 \\    
    \hline
\end{tabular}
\end{center}
\label{tab:real_main_results}
\end{table}

\begin{table}[ht]
\caption{Measured and reconstructed baseline lengths. Five variants of the affine transformation estimation, proposed in Section~\ref{sec:estimation}, are run. Distances given in millimeters. Best values are highlighted by bold font.}
\begin{center}
\begin{tabular}{c c | c  c c c c c c }
\multicolumn{2}{c}{~} & \textbf{Measured} & \multicolumn{5}{c}{\textbf{Reconstructed}} \\
\multicolumn{3}{c}{~}  & \textbf{F2UDIR} & \textbf{F3UDIR} & \textbf{DET3UDIR} & \textbf{2SDIR} & \textbf{3SDIR}\\
\multirow{2}{*}{\textbf{Regular}}  & Std. Stereo & 300 & 292.90 & 289.35 & 293.46 & 293.44 & \textbf{290.70} \\
\multirow{2}{*}{} & Forw. Motion & 150 & 142.81 & 142.54 & 147.62 & 146.29 & \textbf{148.93} \\
\hline
\multirow{2}{*}{\textbf{Fisheye}}  & Std. Stereo & 300 & \textbf{307.10} & 309.22 & 308.91 & 308.66 & 308.62\\
\multirow{2}{*}{} & Forw. Motion & 200 & \textbf{180.28} & 179.65 & 179.5 & 179.77 & 179.81 \\
\hline
\end{tabular}
\end{center}
\label{tab:real_distances}
\end{table}

\section{Conclusion and Future Work}
\label{sec:conclusion}
This paper mainly deals with two important issues: (i) The basic theoretical rules are overviewed for AC-based 3D vision, including basic estimation algorithms; (ii) and it is evaluated how accurate affine transformations are required in order to get high-quality oriented 3D point cloud as output of the proposed 3D reconstruction approach.
We proposed a novel 3D reconstruction pipeline for the validation that exploits affine transformations as well as point locations. The affine transformations are estimated in a novel way contrary to the state-of-the-art methods: they are estimated based on directions. Several types of affine estimators are introduced here, utilizing scaled and unscaled directions. Fundamental matrix of stereo pair can also be exploited for the estimation.

The quantitative evaluation shows that accurate oriented point clouds can be achieved if the noise in the directions are less than a few degrees. Therefore, we think the methods can be applied in realistic cases. However, there are special motion types and planes for which the reconstructions are very sensitive to the noise. These cases are also discussed in the paper.

\noindent \textbf{Future work.} The directions for affine transformation estimation are retrieved from chessboard patterns in our tests. To our opinion, the most important future work is to develop novel estimators that should be a combination of point and direction (line) matches. Accurate retrieve of scales is also an open question. Then we plan to apply the proposed pipeline for processing image pairs of vehicle-mounted cameras. Surface normals are very useful for visual system of autonomous vehicles, as there are many flat surfaces in human-made environment, and they can be detected from surface normals.

In summary, we think that AC-based 3D reconstruction is a new technology that can be utilized in order to make computer vision algorithm more accurate, more robust, therefore, make results more realistic.

\begin{acknowledgements}
Application Domain Specific Highly Reliable IT Solutions” project  has been implemented with the support provided from the National Research, Development and Innovation Fund of Hungary, financed under the Thematic Excellence Programme no. 2020-4.1.1.-TKP2020 (National Challenges Subprogramme) funding scheme.

\noindent The author would like to thank Tekla Tóth and Dániel Baráth for their valuable comments.
\end{acknowledgements}

%
%

\bibliographystyle{spbasic}      
\bibliography{Hajder_affine}   

%
%

\appendix
\section{Estimation of an Affine Transformation with Known Scale}
\label{sec:app}
\noindent \textbf{Problem Statement.} 
There is an inhomogeneous linear system of equations $\mathbf A \mathbf x=0$ to minimize, where
\begin{equation}
\mathbf x=\left[ a_1 ~ a_2 ~ a_3 ~ a_4  ~ \alpha_1 ~ \alpha_2 ~ \cdots ~ \alpha_N \right]^T,
\end{equation}
and there is a non-linear constraint for the vector of unknown variables: $a_{1}a_{4}-a_{2}a_{3}=s$. Regarding the discussed estimation problem in Sec.~\ref{sec:estimation}, $s$ is the known determinant of affine transformation. The aim is to minimize the $L_2$ norm of the error vector $\mathbf A \mathbf x$.

\noindent \textbf{Least Squares Solution.} 
The constrained linear homogeneous problem can be solved by introducing a Lagrange multiplier $\lambda$. The least-squares (LSQ) cost function, denoted by $J$, is as follows:
\begin{equation}
J= {\mathbf x}^{T} {\mathbf A}^{T} \mathbf{A}{\mathbf x}+\lambda\left(a_{1}a_{4}-a_{2}a_{3}-s\right).
\end{equation}
The optimal solution, in the LSQ sense, is obtained by taking the gradient of the cost:
\begin{eqnarray}
\frac{\partial J}{\partial \mathbf x}=2{\mathbf A}^{T} \mathbf{A}{\mathbf x}+\lambda\left[\begin{array}{ccccccc}
a_{4} & -a_{3} & -a_{2} & a_{1} & 0 & \dots & 0 \end{array}\right]^T  = & \\
 2 {\mathbf A}^{T} \mathbf{A}{\mathbf x}+\lambda  \mathbf{B}{\mathbf x}  = & 0,
\end{eqnarray}

where

\begin{equation}
\mathbf B = \left[\begin{array}{ccccccc}
0 & 0 & 0 & 1 & 0 & \cdots & 0\\
0 & 0 & -1 & 0 & 0 & \cdots & 0\\
0 & -1 & 0 & 0 & 0 & \cdots & 0\\
1 & 0 & 0 & 0 & 0 & \cdots & 0\\
0 & 0 & 0 & 0 & 0 & \cdots & 0\\
\vdots & \vdots & \vdots & \vdots & \vdots & \ddots & \vdots\\
0 & 0 & 0 & 0 & 0 & \cdots & 0
\end{array}\right].
\end{equation}

The optimal solution is given via the following generalized eigenvalue problem:
$    {\mathbf A}^{T} \mathbf{A}{\mathbf x}=-\frac{\lambda}{2} \mathbf{B}{\mathbf x}$.
The results can be substituted into the original LSQ cost:
$    {\mathbf x}^T {\mathbf A}^{T} \mathbf{A}{\mathbf x}= - \frac{\lambda}{2} {\mathbf x}^T \mathbf{B}{\mathbf x} = -\lambda s$,
as ${\mathbf x}^T \mathbf{B}{\mathbf x}=2s$. Therefore, the solution is the generalized eigenvector, corresponding to the largest/least\footnote{Largest one should be selected if $s>0$.} generalized eigenvalue.  However, not all generalized eigenvector can be selected: if the signs of $s$ and $a_1 a_4 - a_2 a_3$ are not same, the length of the eigenvector cannot be scaled to the desired value. Therefore, that eigenvalues/vectors must be discarded.

\end{document}